%% file: main.tex
\documentclass[conference]{IEEEtran}

\usepackage{times}
\usepackage[numbers]{natbib}
\IEEEoverridecommandlockouts
\usepackage{multicol}
\usepackage{booktabs}
\usepackage{amsmath}
\usepackage{amssymb}
\usepackage{graphicx}
\usepackage{subcaption}
\usepackage{multirow}
\usepackage[table]{xcolor}
\usepackage[hyphens]{url}
\definecolor{linkblue}{RGB}{35,105,155}
\usepackage{algorithm}
\usepackage{algorithmic}
\usepackage{newfloat}
\usepackage{listings}

\DeclareCaptionStyle{ruled}{
  labelfont=normalfont,
  labelsep=colon,
  strut=off
}
\floatstyle{ruled}
\newfloat{listing}{tb}{lst}{}
\floatname{listing}{Listing}

\usepackage[bookmarks=true,hidelinks]{hyperref}

\InputIfFileExists{preamble.tex}{\input{preamble}}{}
\InputIfFileExists{math.tex}{\input{math}}{}

\begin{document}

\title{PFM-HR: Pose Flow Matching for Humanoid Robots}

\author{
\IEEEauthorblockN{
\textbf{Yukang Gao}\textsuperscript{1,2},
\textbf{Yi Gu}\textsuperscript{1},
\textbf{Yangchen Zhou}\textsuperscript{1},
\textbf{Xingyu Chen}\textsuperscript{1,2},
\textbf{Zhaorui Wang}\textsuperscript{1},
\textbf{Fanghai Zhang}\textsuperscript{1},\\
\textbf{Hanyang Cao}\textsuperscript{1,2},
\textbf{Zhengyang Shen}\textsuperscript{4},
\textbf{Ji Ma}\textsuperscript{2},
\textbf{Runhan Zhang}\textsuperscript{2,3},
\textbf{Lei Han}\textsuperscript{2},
\textbf{Renjing Xu}\textsuperscript{1,\(\dagger\)}
}

\IEEEauthorblockA{
\textsuperscript{1}HKUST(GZ) \quad
\textsuperscript{2}Noitom Robotics \quad
\textsuperscript{3}SIGS, Tsinghua University \quad
\textsuperscript{4}Google\\[0.4em]
\textsuperscript{\(\dagger\)}Corresponding authors
}
}

\IEEEaftertitletext{%
    \vspace{-0.8cm}
    \begin{center}
        \href{https://gaoyukang33.github.io/PFM-HR.web/}{%
            \textcolor{linkblue}{%
                \large\bfseries PFM-HR Project Page
            }%
        }
    \end{center}
}

\maketitle

\begin{abstract}
Motion priors improve reinforcement learning for physics-based humanoid
tracking, but temporal priors require ordered motion clips, while pose priors
provide limited guidance for policy-induced pose transitions. We present
\emph{Pose Flow Matching for Humanoid Robots} (PFM-HR), a reusable flow matching
prior trained directly on large-scale unordered pose data. PFM-HR introduces the
\emph{Pose Geometry Score} (PGS), which quantifies how joint coordinate changes
during rollouts align with the local geometry of pose variation captured by the
prior. Using PGS to modulate the tracking reward guides policy exploration
toward structured pose changes while keeping the prior frozen across tracking
tasks. Experiments demonstrate that PFM-HR improves both single-motion and
general motion tracking, especially for highly dynamic motions.
\end{abstract}


\IEEEpeerreviewmaketitle

\input{Section/Intro}

\input{Section/Relatedwork}

\input{Section/Method}
\input{Section/Experiment}

\section{Conclusion}
\label{sec:conclusion}

We presented PFM-HR, a reusable Flow Matching pose prior for physics-based
humanoid motion tracking that can be trained directly on large-scale unordered
pose data, without pose-distance supervision or temporally segmented motion
clips. Its Pose Geometry Score uses the denoiser Jacobian to evaluate whether
policy-induced pose changes align with the joint co-variation patterns captured
by the prior. This score can be computed efficiently through a
Jacobian--vector product and used to modulate tracking rewards while keeping
the prior frozen. By separating prior learning from policy optimization, PFM-HR
allows a single pretrained model to provide geometry-aware guidance across
different tracking tasks. Experiments demonstrate improved sample efficiency
and tracking accuracy, particularly for dynamic and acrobatic motions, as well
as compatibility with real-world humanoid deployment. These results show that
the local geometry learned from unordered poses can provide useful control
guidance even without explicit temporal supervision.

\section{Limitations}

PFM-HR captures local co-variation in the marginal pose distribution rather
than temporal dynamics. Because the training poses are unordered and PGS is
invariant to direction reversal at a fixed query pose, it cannot assess the
likelihood, direction, or ordering of temporal transitions. Consequently,
opposite transitions may receive similar geometric evaluations even when only
one is consistent with the intended motion. The reliability of PGS also
depends on the coverage of the pose dataset, and its guidance may be less
informative for poses far from the training distribution. Future work will
explore sign-sensitive scores and temporally conditioned priors while
preserving the efficiency and reusability of PFM-HR.
\bibliographystyle{plainnat}
\bibliography{reference}

\clearpage
\appendices
\twocolumn[
  \centering
  {\LARGE\bfseries Supplementary Material\par}
  \vspace{1.5em}
]

\input{Supplementary_Material_arxiv}

\end{document}

%% file: Section/Intro.tex
\section{Introduction}
\label{section:intro}

Humanoid robots hold considerable promise for operating in human-centered
environments, where their humanlike morphology allows them to use the same
spaces, tools, and interaction patterns as people
\cite{videomimic,yin2025visualmimic,zhao2025resmimic,he2026ultra}.
Recent advances in reinforcement learning (RL) have enabled increasingly
capable humanoid locomotion and control. Yet dynamic feasibility alone does
not specify how the joints should vary together to produce coordinated
humanlike motion. Even with task and tracking rewards, the policy receives
limited guidance on such joint dependencies and must discover them through
exploration in a high-dimensional control space. This burden becomes
particularly severe for dynamic motions that require precise whole-body
coordination.

\input{Figures/teaser}

Data-driven motion priors address this problem by guiding policy optimization
toward behaviors supported by demonstrations. DeepMimic tracks reference
trajectories to reproduce complex skills~\cite{peng2018deepmimic}, while
adversarial methods such as AMP and ADD learn discriminator-based rewards that
encourage realistic motion and reduce manual reward engineering
\cite{peng2021amp,escontrela2022adversarial,zhang2025physics}. Although
effective, these discriminator-based priors are commonly trained together with
the control policy, which limits their reuse across tasks. Recent frozen priors
improve modularity but introduce a different trade-off. PDF-HR learns a
reusable pose-distance field from retargeted humanoid data and improves
downstream tracking~\cite{gu2026hr}, but it  requires precomputed
pose--distance supervision and scores each pose without considering the
relative joint changes that produced it. SMP instead studies reusable score-matching motion priors in a different control formulation, repurposing a pretrained motion diffusion model as a reward through score distillation~\cite{mu2026smp}. This captures temporal motion structure,
but requires ordered motion clips and incurs repeated score-model evaluations
during RL. This leaves a gap between efficient pose-wise priors,
which do not evaluate how joints change together, and temporal priors, which
require ordered data and heavier online inference.

We address this gap with \emph{Pose Flow Matching for Humanoid Robots}
(PFM-HR), a reusable Flow Matching prior trained directly on individual
humanoid poses, with the training corpus treated as an unordered
collection~\cite{lipman2023flow}. PFM-HR requires neither pose--distance
supervision nor temporally segmented motion clips. It is pretrained
independently of the control policy, remains frozen during downstream RL, and
can be extended through continued training as additional pose data become
available.


Our key insight is that a denoiser trained only on marginal pose samples still
captures local pose geometry. Under the Gaussian Flow Matching path, the
Jacobian of the population-optimal denoising map is proportional to the
conditional covariance of clean poses
\cite{manor2023posterior,nadar2026posed}. Its directional response therefore
reflects conditional modes of joint co-variation. 

PFM-HR applies this Jacobian to the
normalized joint-coordinate change between consecutive rollout poses and
defines the squared Jacobian--vector response as the
\emph{Pose Geometry Score} (PGS). A high PGS indicates that the relative joint
changes align with co-variation modes strongly represented by the pose prior. After reference-specific calibration, PGS
modulates the tracking reward to reduce learning signals from weakly supported
change patterns. 

Our contributions are threefold:
\begin{itemize}
    \item We introduce a scalable, reusable Flow Matching pose prior trained on
    unordered humanoid poses. We demonstrate pretraining on 60 million poses and efficient expansion through continued training.

    \item We establish a conditional covariance interpretation of the
    denoiser's response and formulate the Pose Geometry Score for
    measuring response along the joint coordinate change patterns of policy generated pose updates.

    \item We integrate PGS into an efficient, reference calibrated reward
    modulation mechanism for humanoid tracking. Experiments on
    single-trajectory tracking and general motion tracking benchmarks demonstrate improved sample
    efficiency and tracking accuracy over baselines.
\end{itemize}

%% file: Figures/teaser.tex
\begin{figure*}[t]
\centering
\includegraphics[width=0.99\linewidth]{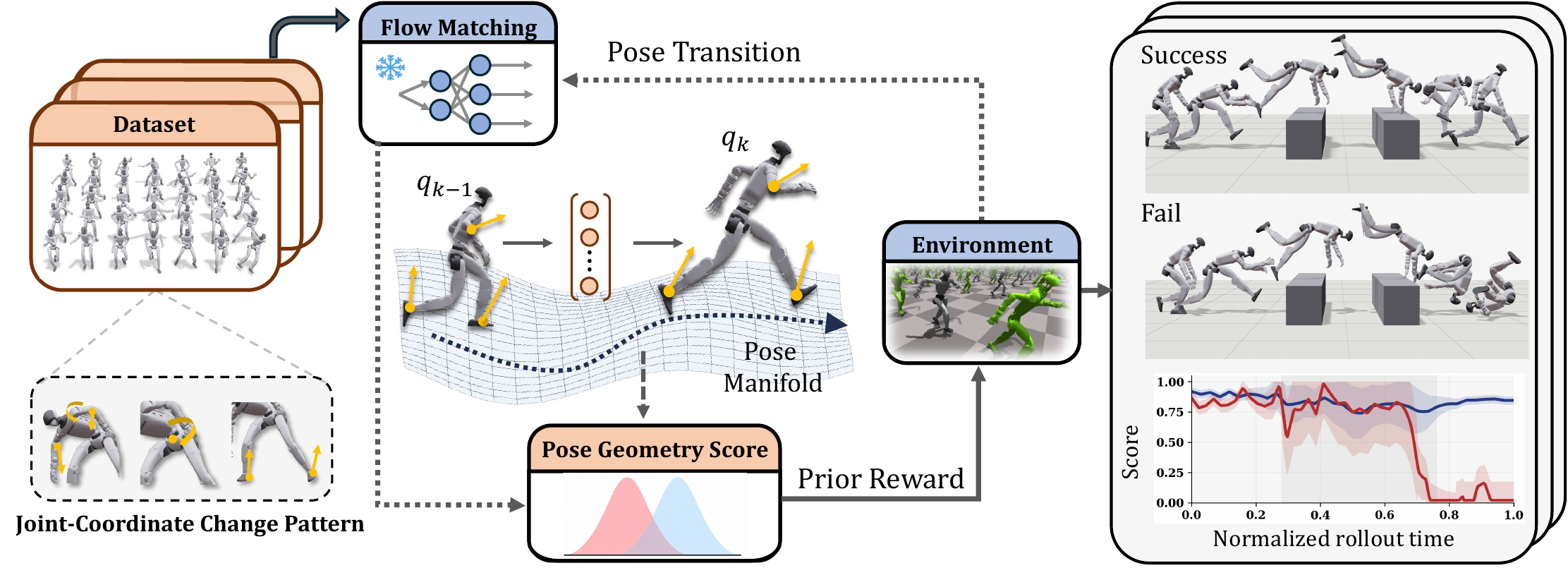}

\caption{Overview of PFM-HR.
PFM-HR learns a reusable Flow Matching pose prior from large-scale, unordered
pose data and keeps it frozen during policy learning. The yellow arrows
illustrate representative joint-coordinate changes between consecutive rollout
poses, whose normalized pattern is evaluated by the \emph{Pose Geometry Score}  (PGS)
through the prior's local response. PGS then modulates the tracking reward to
guide dynamic motion tracking.}

\label{fig:teaser}
\end{figure*}

%% file: Section/Relatedwork.tex
\section{Related Work}

\noindent
\textbf{Physics-based Humanoid Motion Tracking.}
Deep reinforcement learning has become a central paradigm for physics-based
humanoid motion tracking. DeepMimic~\cite{peng2018deepmimic} demonstrated that
physically simulated characters can reproduce highly dynamic skills by
tracking motion-capture references. Subsequent work learns reusable motion
representations to improve control and downstream task learning. These include
VAE-based controllers
\cite{controlvae22,PhysicsVAE,tessler2024maskedmimic},
discrete representations~\cite{yao2024moconvq,zhu2023neural}, and reusable
skill spaces such as ASE, CALM, and PULSE
\cite{peng2022ase,tessler2023calm,luo2024universal}. Adversarial methods instead
derive motion-style rewards directly from demonstrations. AMP
\cite{peng2021amp,escontrela2022adversarial} replaces hand-designed style
objectives with a discriminator, while ADD~\cite{zhang2025physics} extends this
formulation by balancing multiple tracking objectives.

Large-scale trackers further improve motion coverage and robustness.
PHC~\cite{luo2023perpetual} scales physics-based imitation to thousands of motion
clips and supports recovery from large deviations. More recent systems,
including GMT~\cite{chen2025gmt},
BeyondMimic~\cite{liao2025beyondmimicmotiontrackingversatile}, and
SONIC~\cite{luo2025sonic}, target diverse and previously unseen motions. Heracles~\cite{tao2026heracles} further connects
generative motion synthesis with physics-based tracking through a
state-conditioned middleware that adapts high-level references under large
state deviations. These methods substantially broaden
the capabilities of humanoid controllers, but their motion knowledge is
typically embedded in a pretrained policy or learned jointly with the
controller. Scaling them therefore requires large motion datasets
\cite{mahmood2019amass,harvey2020robust,bones2026ai} and extensive policy
training. PFM-HR instead studies a frozen, controller-independent prior that can
be attached to task-specific tracking policies as a dense reward signal.


\noindent
\textbf{Human and Humanoid Motion Priors.}
Human pose and motion priors are widely used to regularize motion recovery and
synthesis
\cite{lu2024dposerdiffusionmodelrobust,lu2025dposerx,zhang2024rohm}.
PoseNDF~\cite{tiwari2022pose} models pose plausibility with an implicit neural
distance field. NRDF~\cite{he24nrdf} improves its geometric consistency, while
NRMF~\cite{yu2025nrmf} extends distance-based priors to continuous motion.
PDF-HR~\cite{gu2026hr} adapts neural pose distance fields to humanoid
control and demonstrates that a frozen pose prior can improve downstream
tracking. These methods provide reusable state-level regularization, but
constructing distance supervision requires an additional sampling stage.

Generative models provide another route to reusable priors.
SMP~\cite{mu2026smp} converts a pretrained motion diffusion model into an RL
reward through Score Distillation Sampling, separating prior learning from
policy optimization while retaining temporal motion information. This
motion-level representation captures dependencies across consecutive states,
but requires ordered clips. In contrast, priors trained on individual poses can use unordered
pose collections and avoid modeling fixed temporal windows. However, they
typically evaluate each pose independently and do not explicitly characterize
how joint coordinates vary together around the current pose.

This gap motivates a reusable prior that can be trained on unordered pose
collections while still revealing how joint coordinates vary together. Such a
prior would retain the efficiency of pose based modeling while providing richer
guidance than an independent pose plausibility score.

%% file: Section/Method.tex
\section{Methodology}

\subsection{Problem Formulation}
\label{sec:overview}

\noindent
\textbf{Robot pose representation.}
We represent a humanoid pose by its actuated joint configuration. Let the robot
have $N_J$ actuated degrees of freedom. At control step $k$, the pose is written
as~\cite{gu2026hr}
\begin{equation}
\mathbf{q}_k
=
[q_k^1, q_k^2, \dots, q_k^{N_J}]^\top
\in
\mathcal{Q}
\subset
\mathbb{R}^{N_J},
\label{eq:state_def}
\end{equation}
where $q_k^i$ denotes the $i$-th actuated joint coordinate and
$\mathcal{Q}$ is bounded by the robot joint limits. We use normalized joint
coordinates as the input representation for the pose prior. For brevity,
$\mathbf{q}_k$ denotes the normalized pose hereafter.

During a rollout, the policy generates a pose sequence
$\{\mathbf{q}_k\}_{k=0}^{T}$. We characterize the rollout update ending at
step $k$ by the finite-difference joint-coordinate velocity
\begin{equation}
\label{eq:joint_transition_q}
\boldsymbol{\xi}_{\mathrm{env},k}
=
\frac{
\mathbf{q}_k-\mathbf{q}_{k-1}
}{
\Delta t
}
\in
\mathbb{R}^{N_J},
\end{equation}
where $\Delta t$ is the control interval. While $\mathbf{q}_{k-1}$ identifies
the location of the rollout in pose space,
$\boldsymbol{\xi}_{\mathrm{env},k}$ records the joint-coordinate changes
induced by the policy over one control interval. This pattern describes which joint coordinates
vary together, including their relative magnitudes and signs.

\noindent
\textbf{RL-based motion tracking.}
We formulate humanoid motion tracking as a finite-horizon discrete-time Markov
decision process
$\mathcal{M}
=
\langle\mathcal{S},\mathcal{A},\mathcal{P},r,\gamma\rangle$.
At each control step $k$, the policy $\pi_\theta$ observes
$\mathbf{s}_k\in\mathcal{S}$ and samples an action
$\mathbf{a}_k \sim \pi_\theta(\cdot\mid\mathbf{s}_k)$. The simulator then
evolves according to
$\mathbf{s}_{k+1}\sim
\mathcal{P}(\cdot\mid\mathbf{s}_k,\mathbf{a}_k)$.
The policy is trained to maximize
\begin{equation}
J(\pi_\theta)
=
\mathbb{E}_{\tau\sim\pi_\theta}
\left[
\sum_{k=0}^{T-1}
\gamma^k r(\mathbf{s}_k,\mathbf{a}_k)
\right].
\label{eq:rl_objective}
\end{equation}

In humanoid motion tracking, the reward commonly combines task and tracking
terms,
\begin{equation}
r_k
=
w^{G} r_k^{G}
+
w^{T} r_k^{T},
\end{equation}
where $r_k^{G}$ measures task progress, $r_k^{T}$ measures agreement with the
reference motion, and $w^{G}$ and $w^{T}$ are scalar weights. 
Optimizing these
task-specific objectives alone provides no explicit guidance from the
statistical dependencies among joint coordinates in human pose data.

PFM-HR augments the tracking reward using a frozen Flow Matching pose prior
trained on unordered poses. At each rollout step, PGS applies the denoiser
Jacobian at $\mathbf{q}_{k-1}$ to the normalized joint-coordinate change,
measuring its alignment with the joint co-variation structure encoded by the
pose distribution.

\subsection{Pose Prior Learning with Flow Matching}
\label{sec:pose_prior}

We pretrain an unconditional Flow Matching model on individual poses sampled
from a large-scale motion corpus. The training data are treated as an unordered
pose collection, from which the model learns the marginal pose distribution
$p_{\mathrm{data}}$. Let $\boldsymbol{x}\sim p_{\mathrm{data}}$ be a clean pose and
$\boldsymbol{\epsilon}\sim\mathcal{N}(\boldsymbol{0},\mathbf{I})$ an
independent Gaussian sample. Both training poses $\boldsymbol{x}$ and rollout
poses $\boldsymbol{q}_k$ use the normalized representation defined in
Section~\ref{sec:overview}. We sample $t$ from a logit-normal distribution and construct the linear Flow Matching
path~\cite{lipman2023flow,liu2022flow}
\begin{equation}
\boldsymbol{z}_t
=
t\boldsymbol{x}
+
(1-t)\boldsymbol{\epsilon}.
\label{eq:FM_path}
\end{equation}
This path interpolates between the Gaussian source distribution at $t=0$ and
the pose distribution at $t=1$. Its conditional velocity is
\begin{equation}
\boldsymbol{v}
=
\frac{\mathrm{d}\boldsymbol{z}_t}{\mathrm{d}t}
=
\boldsymbol{x}-\boldsymbol{\epsilon}.
\label{eq:FM_velocity}
\end{equation}

Following JiT~\cite{li2025back}, we parameterize the velocity field through a
clean-pose predictor
$\hat{\boldsymbol{x}}_\phi(\boldsymbol{z}_t,t)$. The predicted velocity is
\begin{equation}
\boldsymbol{v}_\phi(\boldsymbol{z}_t,t)
=
\frac{
\hat{\boldsymbol{x}}_\phi(\boldsymbol{z}_t,t)
-
\boldsymbol{z}_t
}{
\max(1-t,\delta)
},
\label{eq:predicted_velocity}
\end{equation}
where $\delta>0$ stabilizes the denominator near $t=1$. We optimize the
Flow Matching objective
\begin{equation}
\mathcal{L}_{\mathrm{FM}}(\phi)
=
\mathbb{E}_{
t,\boldsymbol{x},\boldsymbol{\epsilon}
}
\left[
\left\|
\boldsymbol{v}
-
\boldsymbol{v}_\phi(\boldsymbol{z}_t,t)
\right\|_2^2
\right].
\label{eq:FM_objective}
\end{equation}

For $t\leq1-\delta$, the velocity residual satisfies
$\boldsymbol{v}-\boldsymbol{v}_\phi
=
(\boldsymbol{x}-\hat{\boldsymbol{x}}_\phi)/(1-t)$.
Therefore, at any fixed timestep in this regime, minimizing
Eq.~\eqref{eq:FM_objective} is equivalent to clean pose regression, whose
population optimal predictor is the conditional mean
\begin{equation}
F_t^*(\boldsymbol{z})
=
\mathbb{E}
\left[
\boldsymbol{x}
\mid
\boldsymbol{z}_t=\boldsymbol{z}
\right].
\label{eq:population_denoiser}
\end{equation}

We denote the learned approximation to this conditional mean by
\begin{equation}
F_\phi(\boldsymbol{z},t)
:=
\hat{\boldsymbol{x}}_\phi(\boldsymbol{z},t).
\label{eq:denoise_map}
\end{equation}
During policy learning, $F_\phi$ remains frozen and is queried at
$(\boldsymbol{q}_{k-1},t_{\mathrm{eval}})$, where
$t_{\mathrm{eval}}\leq 1-\delta$.

\subsection{Pose Geometry Score}
\label{sec:transition_response}

The Pose Geometry Score measures whether the relative joint changes produced by the policy agree
with the conditional co-variation structure encoded by the pose prior. 

\noindent
\textbf{Denoiser-induced pose geometry.}
For fixed $t\leq1-\delta$, the population-optimal denoiser in
Eq.~\eqref{eq:population_denoiser} satisfies~\cite{manor2023posterior}
\begin{equation}
\mathbf{J}_t^\star(\boldsymbol{z})
:=
\nabla_{\boldsymbol{z}}F_t^\star(\boldsymbol{z})
=
\frac{t}{(1-t)^2}\mathbf{C}_t(\boldsymbol{z}),
\label{eq:jacobian_covariance}
\end{equation}
where $\mathbf{C}_t(\boldsymbol{z})
=\operatorname{Cov}[\boldsymbol{x}\mid
\boldsymbol{z}_t=\boldsymbol{z}]$. In particular, its off-diagonal entries
describe conditional co-variation between joint coordinates. For the learned
denoiser, we define $\mathbf{J}_\phi=\nabla_{\boldsymbol{z}}F_\phi$ and
\begin{equation}
\mathbf{G}_\phi(\boldsymbol{z},t)
=
\mathbf{J}_\phi(\boldsymbol{z},t)^\top
\mathbf{J}_\phi(\boldsymbol{z},t)
\succeq\mathbf{0}.
\label{eq:denoiser_pose_geometry}
\end{equation}

\noindent
\textbf{Pose Geometry Score.}
At rollout step $k$, we construct a query using the same corruption
parameterization as in pretraining:
\begin{equation}
\widetilde{\mathbf{q}}_{k-1}
=
t_{\mathrm{eval}}\mathbf{q}_{k-1}
+
(1-t_{\mathrm{eval}})\boldsymbol{\epsilon}_k,
\quad
\boldsymbol{\epsilon}_k\sim
\mathcal{N}(\mathbf{0},\mathbf{I}),
\label{eq:rollout_noisy_query}
\end{equation}
where $t_{\mathrm{eval}}\leq1-\delta$ is fixed during policy learning. We then
set
$\mathbf{J}_{\phi,k}
=\mathbf{J}_\phi(\widetilde{\mathbf{q}}_{k-1},t_{\mathrm{eval}})$ and
$\mathbf{G}_{\phi,k}
=\mathbf{J}_{\phi,k}^{\top}\mathbf{J}_{\phi,k}$.

\input{Figures/radar_charts}

The joint-coordinate change pattern is represented by
\begin{equation}
\boldsymbol{d}_k
=
\frac{\boldsymbol{\xi}_{\mathrm{env},k}}
{\sqrt{\|\boldsymbol{\xi}_{\mathrm{env},k}\|_2^2+
\varepsilon_{\mathrm{num}}}},
\label{eq:normalized_rollout_direction}
\end{equation}
which retains the relative changes across joints while reducing sensitivity to
their overall magnitude. We define the \emph{Pose Geometry Score} (PGS) as
\begin{equation}
s_{\mathrm{PGS},k}
:=
\|\mathbf{J}_{\phi,k}\boldsymbol{d}_k\|_2^2
=
\frac{\|\mathbf{J}_{\phi,k}
\boldsymbol{\xi}_{\mathrm{env},k}\|_2^2}
{\|\boldsymbol{\xi}_{\mathrm{env},k}\|_2^2+
\varepsilon_{\mathrm{num}}}.
\label{eq:pose_geometry_score}
\end{equation}
The score is computed by a Jacobian--vector product without explicitly
constructing the Jacobian or geometric tensor.

At the population optimum, with
$\mathbf{C}_k
=\mathbf{C}_{t_{\mathrm{eval}}}(\widetilde{\mathbf{q}}_{k-1})$,
\begin{equation}
s_{\mathrm{PGS},k}^{\star}
=
\left(
\frac{t_{\mathrm{eval}}}
{(1-t_{\mathrm{eval}})^2}
\right)^2
\boldsymbol{d}_k^\top\mathbf{C}_k^2\boldsymbol{d}_k.
\label{eq:population_pose_geometry_score}
\end{equation}
This expression weights the projection of $\boldsymbol{d}_k$ onto each
conditional co-variation mode by its squared variance. A high PGS therefore
indicates a joint-change pattern strongly supported by the pose prior. After
calibration, PGS increases the reward contribution of such coordinated updates
and reduces that of weakly supported patterns, biasing policy optimization
toward coordinated humanoid poses. The detailed covariance derivation is provided in the supplementary material.

\subsection{Score Calibration and Reward Modulation}
\label{sec:reward_modulation}

\textbf{Evaluation-Timestep Selection.}
The evaluation timestep determines the noise scale at which the prior is
queried and therefore affects the sensitivity of pose geometry score to pose-change patterns.
We examine this sensitivity under the general motion tracking setting. For each
$t_{\mathrm{eval}}\in
\{0.4,0.5,0.6,0.7,0.75,0.8,0.9\}$, we train a single policy on the same
34-sequence subset of LaFAN1~\cite{harvey2020robust}, excluding the six
\emph{FallAndGetUp} sequences. Each policy is trained for 12 billion samples
and evaluated with tracking horizons of 10, 20, and 30 seconds, while all other
settings are held fixed.
Figure~\ref{fig:timestep} shows that intermediate timesteps
provide faster convergence and lower position errors. Strong
corruption at small $t$ reduces pose specificity, whereas near
$t=1$ the denoising map approaches the identity and becomes
less discriminative. We therefore use
$t_{\mathrm{eval}}=0.75$ throughout the remaining experiments. Additional timestep selection details are included in the supplementary material.

\noindent
\textbf{Reference-calibrated reward modulation.}
For each transition in a reference motion $\mathcal{R}_\tau$, we define
$\boldsymbol{\xi}^{\mathrm{ref}}_n =
(\boldsymbol{q}^{\mathrm{ref}}_n-
\boldsymbol{q}^{\mathrm{ref}}_{n-1})/\Delta t$
and obtain $\boldsymbol{d}^{\mathrm{ref}}_n$ using
Eq.~\eqref{eq:normalized_rollout_direction}. Before policy training,
we compute PGS for all reference transitions and construct the empirical
CDF $\widehat{F}_\tau$. During rollout, the score is mapped to its
reference percentile $u_k^\tau=\widehat{F}_\tau(s_{\mathrm{PGS},k})$
and converted into
\begin{equation}
\rho_k^\tau
=
\operatorname{clip}
\left(
\frac{p_{\mathrm{good}}-u_k^\tau}
{p_{\mathrm{good}}-p_{\mathrm{bad}}},
0,1
\right),
\qquad
r_k^P=\exp(-\alpha\rho_k^\tau),
\label{eq:pgs_percentile_calibration}
\end{equation}
where $\alpha=0.5$, $p_{\mathrm{good}}=0.05$, and
$p_{\mathrm{bad}}=0.01$. The final reward is
\begin{equation}
r_k=w^G r_k^G+w^T r_k^T r_k^P.
\label{eq:final_reward}
\end{equation}
The tracking reward is unchanged above $p_{\mathrm{good}}$ and
maximally attenuated below $p_{\mathrm{bad}}$.

\input{Figures/timestep}

%% file: Figures/radar_charts.tex
\begin{figure}[t]
\centering
\includegraphics[width=1.0\linewidth]{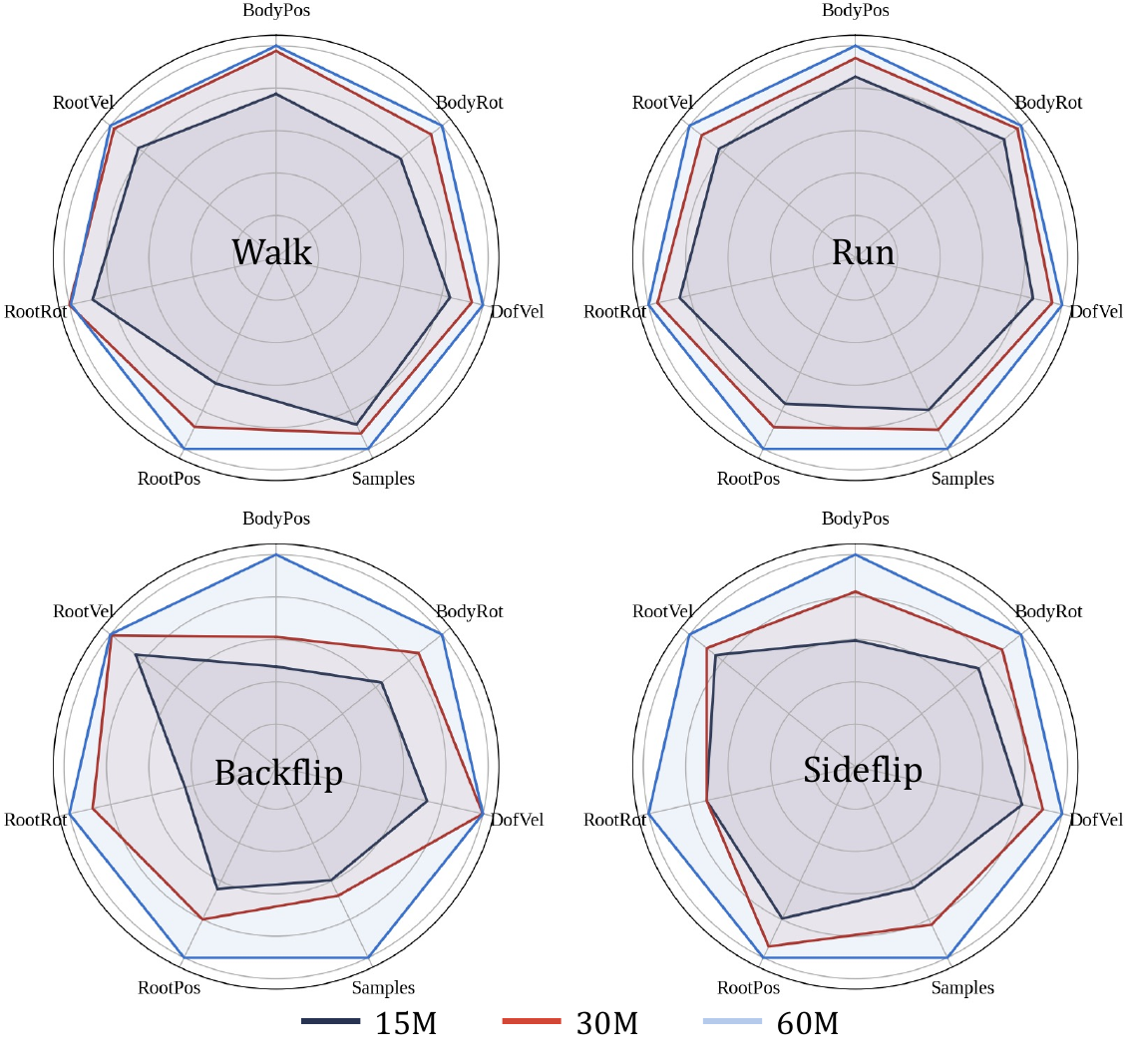}
\caption{Impact of prior-training data scale.
Tracking performance across four representative motions using nested \(15\)M, \(30\)M,
and \(60\)M pose corpora. Metrics are min-normalized, with larger values
indicating lower errors.}
\label{fig:radar_chart} 

\end{figure}

%% file: Figures/timestep.tex
\begin{figure}[h]
\centering
\includegraphics[width=0.99\linewidth]{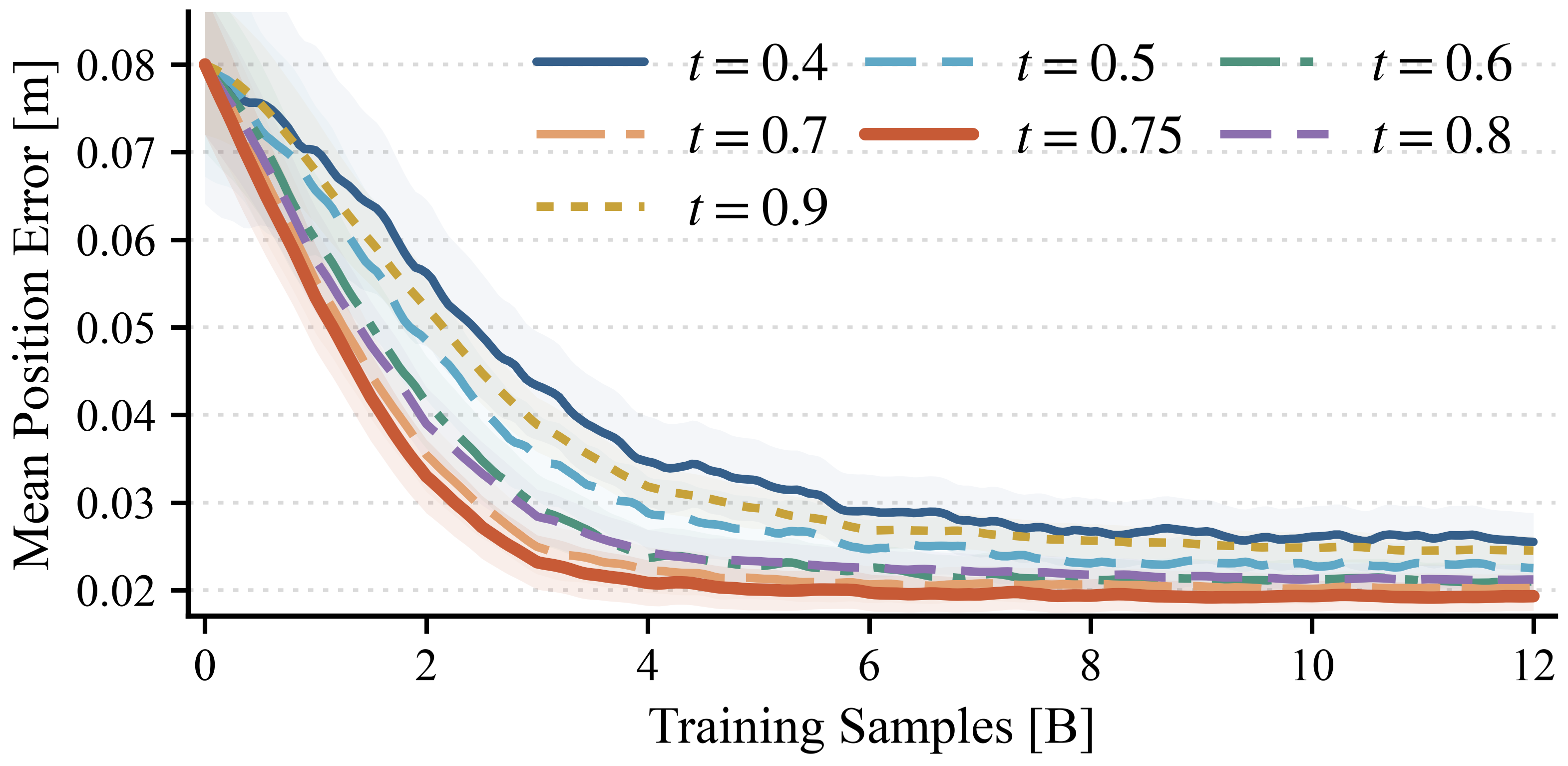}
\caption{Sensitivity to the PGS evaluation timestep in general motion
tracking. We report the position error during policy training, averaged over
the 10, 20, and 30\,s evaluation horizons, for different
$t_{\mathrm{eval}}$. Curves and shaded regions denote the
mean and $\pm1$ standard deviation over three seeds.}

\label{fig:timestep}
\end{figure}

%% file: Section/Experiment.tex
\section{Experiments}
\subsection{Experimental Setup}
\noindent
\textbf{Baselines and prior training.} Across both tracking settings, we compare vanilla
ADD~\cite{zhang2025physics}, ADD w/ PDF-HR~\cite{gu2026hr}, and ADD w/
PFM-HR under the same ADD-based motion tracking backbone; score-matching prior SMP~\cite{mu2026smp} uses a different control
formulation and is not directly compatible with this plug-in comparison. For a
controlled comparison, PDF-HR and PFM-HR are pretrained on the same unmirrored
BONES-SEED~\cite{bones2026ai} pose corpus. PDF-HR constructs pose--distance
pairs using its official implementation.

\noindent
\textbf{Evaluation metrics.}
We evaluate sample efficiency and tracking accuracy. Policies are evaluated
every 100 training iterations on 4096 test episodes. Success rate is defined as
the mean fraction of reference trajectory tracked continuously. Sample efficiency is measured by the number of training samples
required to first reach an $80\%$ success rate. We additionally measure position
and rotation errors:
\begin{align}
E^{\mathrm{pos}}
&=
\frac{1}{N+1}
\left(
\sum_{j=1}^{N}e_j^{\mathrm{pos}}
+e_{\mathrm{root}}^{\mathrm{pos}}
\right),\\
E^{\mathrm{rot}}
&=
\frac{1}{N+1}
\left(
\sum_{j=1}^{N}e_j^{\mathrm{rot}}
+e_{\mathrm{root}}^{\mathrm{rot}}
\right),
\end{align}
where $N$ is the number of non-root joints. Joint errors are measured in the
robot-local frame, while root errors are measured in the global frame. Lower
values indicate better tracking accuracy. Due to space constraints, additional metrics and rotation errors are reported in the supplementary material.

\noindent
\textbf{Implementation details.}
All policies are trained on 8 NVIDIA RTX 4090 GPUs using 4096 parallel
environments and the default MimicKit configuration. PFM-HR uses a residual
MLP with 10 blocks and a hidden dimension of 1024. The Flow Matching timestep
is injected through adaLN-Zero conditioning~\cite{peebles2023scalable}, and
the pretrained model remains frozen during policy training. Further implementation details for pretraining and policy learning are provided in the supplementary material.

\subsection{Single-trajectory Motion Tracking}
\noindent
\textbf{Experiment settings.}
For single-trajectory tracking, we compare vanilla
ADD~\cite{zhang2025physics} with two prior-augmented variants,
ADD w/ PDF-HR~\cite{gu2026hr} and ADD w/ PFM-HR, on nine MimicKit
tasks~\cite{peng2025mimickit}: \emph{Walk}, \emph{Run}, \emph{Jump},
\emph{Spinkick}, \emph{Cartwheel}, \emph{Backflip}, \emph{Sideflip},
\emph{Double Kong}, and \emph{Speed Vault}.

\noindent
\textbf{Results.} Table~\ref{table:single_motion_tracking_exp} reports sample efficiency and
positional tracking error. ADD w/ PDF-HR is more sample-efficient on \emph{Walk},
\emph{Run}, and \emph{Spinkick}, while ADD w/ PFM-HR performs best on the other
six tasks. Both prior-based methods solve \emph{Backflip} and \emph{Double Kong},
where vanilla ADD fails, with PFM-HR converging faster on both.
ADD w/ PFM-HR achieves the lowest or tied-lowest mean positional error on seven
tasks and lower mean positional errors than PDF-HR on eight, although the
margins are generally small. Overall, PFM-HR combines competitive sample
efficiency with low tracking errors, with larger sample-efficiency gains on
challenging dynamic motions such as \emph{Backflip} and \emph{Double Kong}.
PGS evaluates the prior response along rollout pose changes to guide joint
coordination, while the tracking objective specifies the target motion.
Qualitative comparisons are shown in Fig.~\ref{fig:comparison}.

\input{Figures/teaser_real}

\noindent
\textbf{Real-world deployment.}
We use the BeyondMimic training and deployment
pipeline~\cite{liao2025beyondmimicmotiontrackingversatile}, adding
PFM-HR solely as a frozen prior during simulation training while leaving
the deployment pipeline unchanged.

\input{Table/real_deploy}

We evaluate four dynamic skills: \emph{spinkick, kick combo, cartwheel},
and \emph{backflip}. In Table~\ref{table:real_deploy}, BM denotes the
original BeyondMimic method without an auxiliary pose prior, while
BM w/ PDF-HR and BM w/ PFM-HR augment BM with PDF-HR and our PFM-HR,
respectively. BM w/ PFM-HR requires the fewest simulation samples to reach
$SR \geq 80\%$ across all skills, reducing the requirements by
$15.1\%$--$30.8\%$ relative to BM. The successful executions in
Fig.~\ref{fig:teaser_real} demonstrate PFM-HR's compatibility with
real-world deployment without additional deployment-time computation.

\input{Figures/comparison}

\subsection{General Motion Tracking} 

\noindent
\textbf{Experiment settings.}
For general motion tracking, we train a single policy on a subset of
LaFAN1~\cite{harvey2020robust}. The subset contains 34 sequences, excluding six
\emph{FallAndGetUp} sequences. Each
policy is trained for 12 billion samples and evaluated with tracking horizons
of 10, 20, and 30 seconds. 

\noindent
\textbf{Results.} Table~\ref{table:single_motion_tracking_exp} reports general motion tracking errors across
different episode lengths. Under the same ADD backbone and training budget,
PFM-HR achieves the lowest position and rotation errors at all three horizons.
Averaged across horizons, it reduces position error by $7.6\%$ over ADD and
$10.3\%$ over PDF-HR, while reducing rotation error by $3.6\%$ and $7.3\%$,
respectively. In contrast, PDF-HR yields errors comparable to or higher than
vanilla ADD.

\subsection{Ablation Study}

\noindent
\textbf{Impact of Prior-Data Scale and Coverage.}
We train pose priors on three nested BONES-SEED
subsets~\cite{bones2026ai},
$\mathcal{D}_{15\mathrm{M}}\subset
\mathcal{D}_{30\mathrm{M}}\subset
\mathcal{D}_{60\mathrm{M}}$.
The $15$M subset is balanced across \emph{Locomotion}, \emph{Dances},
\emph{Gaming}, \emph{Sport}, and \emph{Other}; the $30$M subset adds
stratified samples from the same categories; and the $60$M subset contains the
full corpus. All model and policy settings are fixed, and the
reference-specific PGS thresholds are recomputed for each prior. Figure~\ref{fig:radar_chart} shows that larger pretraining sets generally
improve both locomotion and acrobatic tracking, with the $60$M prior achieving
the strongest overall performance. Because scale and coverage increase
together, the results reflect their combined effect.

Extending the $30$M prior to $60$M requires $65$ GPU-hours, compared with
$100$ GPU-hours for training from scratch. At the same scale, generating the
pose-distance supervision for PDF-HR~\cite{gu2026hr} requires over $500$
GPU-hours in our implementation, excluding network training.

\input{Table/ADD_single_motion_tracking}
\input{Table/ablation}

\noindent

\noindent
\textbf{Reward Formulation Comparison.}
We compare ADD, FM-Recon, and PFM-HR to separate the contribution of the
learned prior from its reward formulation. ADD provides a prior-free baseline.
FM-Recon uses the same frozen denoiser to score individual poses through
reconstruction~\cite{yang2025text,zhou2025score}, whereas PFM-HR evaluates
Jacobian response along rollout pose changes. For rollout pose $\boldsymbol{q}_k$, FM-Recon evaluates three noise levels
$\mathcal{T}_{\mathrm{FM}} = \{0.55, 0.65, 0.80\}$, whose signal-to-noise ratios follow the noise-level ensemble used by SMP. At each $t_m$, it samples
$\boldsymbol{z}_{k,m}
=t_m\boldsymbol{q}_k+(1-t_m)\boldsymbol{\epsilon}_{k,m}$ and computes
\begin{equation}
e_{k,m}^{\mathrm{FM}}
=
\left\|
F_\phi(\boldsymbol{z}_{k,m},t_m)
-\boldsymbol{q}_k
\right\|_2^2.
\label{eq:fm_recon_reward}
\end{equation}
To account for scale differences across noise levels, each error is converted
to its reverse percentile under the corresponding reference distribution.
The resulting percentiles are averaged and mapped to the same reward range as
PGS, with all other settings unchanged.

Comparing ADD with FM-Recon evaluates the contribution of the frozen prior,
while comparing FM-Recon with PFM-HR assesses multi-level pose reconstruction
against directional Jacobian scoring. Table~\ref{table:combined_ablation}
shows that PFM-HR provides larger improvements on dynamic motions. With a batch
size of 4096 on an NVIDIA GeForce RTX 4090 GPU, three denoiser evaluations
require $1.8$\,ms, whereas one Jacobian--vector product requires $0.75$\,ms.

\noindent
\textbf{Prediction Parameterization Comparison.}
We compare $\boldsymbol{v}$-prediction and $\boldsymbol{x}$-prediction
while keeping the training data, architecture, Flow Matching path, and
policy-training configuration fixed. For $\boldsymbol{v}$-prediction,
the clean-pose map used by PGS is recovered as
\begin{equation}
F_\phi^{(\boldsymbol{v})}(\boldsymbol{z},t)
=
\boldsymbol{z}
+
(1-t)\boldsymbol{v}_\phi(\boldsymbol{z},t),
\qquad
t\leq 1-\delta.
\label{eq:vpred_denoising_map}
\end{equation}
We then construct PGS from its input Jacobian using the same evaluation
timestep, calibration, and reward modulation, thereby isolating the
effect of the prediction target. As shown in Table~\ref{table:combined_ablation}, both parameterizations
perform similarly on basic locomotion, whereas $\boldsymbol{x}$-prediction
performs better on dynamic motions, particularly \emph{Backflip}.
Because the two variants share the same architecture and prior-query cost,
the results suggest that direct clean-pose prediction provides a more
effective Jacobian response for a  finite-capacity model.

%% file: Figures/teaser_real.tex
\begin{figure}[ht]
\centering
\includegraphics[width=0.99\linewidth]{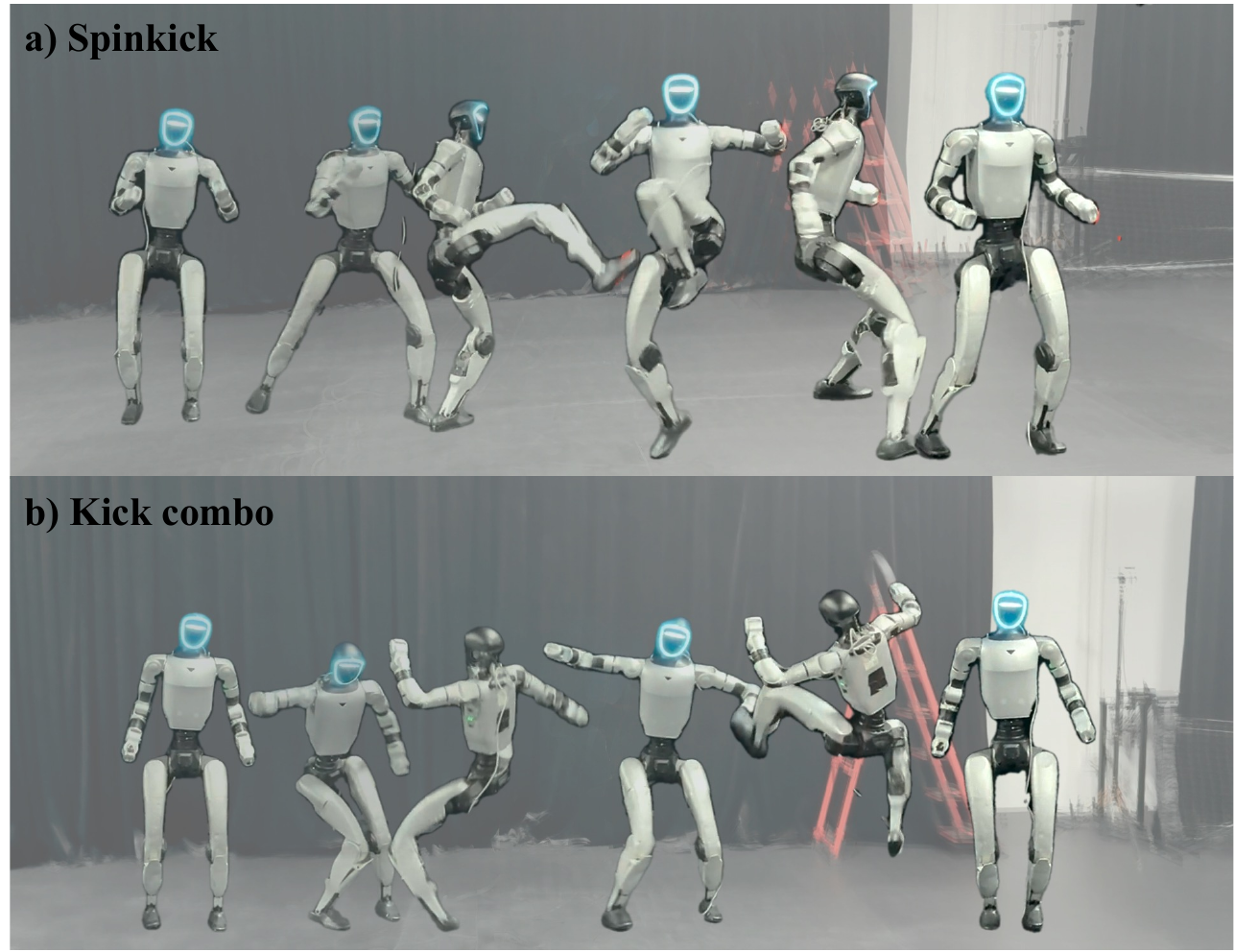}

\caption{Real-world deployment with PFM-HR.
Motion tracking policies trained with PFM-HR execute diverse highly-dynamic
whole-body motions on a humanoid robot.}

\label{fig:teaser_real}
\end{figure}

%% file: Table/real_deploy.tex
\begin{table}[ht]
\centering
\small
\setlength{\tabcolsep}{2pt}
\begin{tabular}{@{}lccc@{}}
\toprule
\textbf{Skill}
& \textbf{BM}
& \textbf{BM w/ PDF-HR}
& \textbf{BM w/ PFM-HR} \\
\midrule
Spinkick
& $331.776^{\pm 18.536}$
& $279.098^{\pm 6.179}$
& $\cellcolor[HTML]{E6E6FD}{\mathbf{251.425}^{\pm 0.000}}$ \\

Kick combo
& $399.515^{\pm 16.348}$
& $347.283^{\pm 6.179}$
& \cellcolor[HTML]{E6E6FD}{$\mathbf{339.110}^{\pm 6.179}$} \\

Cartwheel
& $409.117^{\pm 12.358}$
& $291.742^{\pm 18.536}$
& \cellcolor[HTML]{E6E6FD}{$\mathbf{283.225}^{\pm 10.702}$} \\

Backflip
& $443.052^{\pm 6.179}$
& $318.620^{\pm 12.358}$
& \cellcolor[HTML]{E6E6FD}{$\mathbf{309.794}^{\pm 6.179}$} \\
\bottomrule
\end{tabular}
\caption{Sample efficiency across prior configurations in BeyondMimic.
We report the simulation training samples required to reach
$SR \geq 80\%$ in millions. Lower values are better, and the best
means are highlighted.}
\label{table:real_deploy}
\end{table}

%% file: Figures/comparison.tex
\begin{figure*}[t] 
\centering
\includegraphics[width=0.99\linewidth]{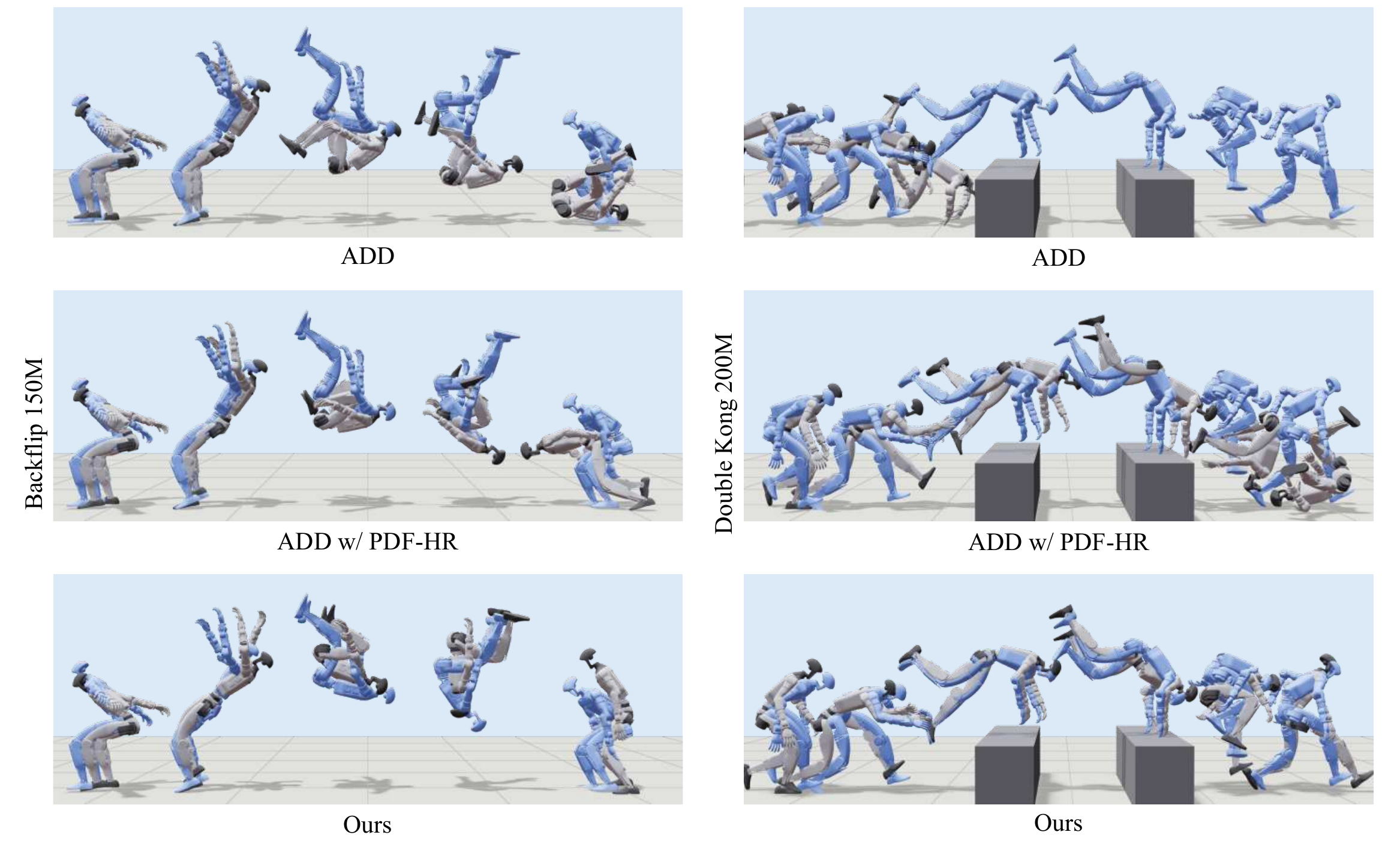}
\caption{ Qualitative comparison of motion tracking performance. The target reference motion is depicted by the blue humanoid, while the executed tracking behaviors produced by different methods are represented by the gray humanoids.}
\label{fig:comparison} 
\end{figure*}

%% file: Table/ADD_single_motion_tracking.tex
\begin{table*}[t]
\centering
\small
\setlength{\tabcolsep}{4pt}
\begin{tabular}{lccc ccc}
\toprule
\multicolumn{7}{c}{\textbf{Single-Trajectory Tracking}} \\
\midrule
\textbf{Skill}
& \multicolumn{3}{c}{\textbf{Samples ($SR \geq 80\%$) [M] ($\downarrow$)}}
& \multicolumn{3}{c}{\textbf{Position Error [m] ($\downarrow$)}} \\
\cmidrule(lr){2-4}
\cmidrule(lr){5-7}
& \textbf{ADD} & \textbf{ADD w/ PDF-HR} & \textbf{ADD w/ PFM-HR}
& \textbf{ADD} & \textbf{ADD w/ PDF-HR} & \textbf{ADD w/ PFM-HR} \\
\midrule
Walk
& $70.036^{\pm 6.179}$
& \cellcolor[HTML]{E6E6FD}{$\mathbf{56.929}^{\pm 12.358}$}
& $65.667^{\pm 0.000}$
& \cellcolor[HTML]{E6E6FD}{$\mathbf{0.008}^{\pm 0.001}$}
& $0.0093^{\pm 0.0011}$
& \cellcolor[HTML]{E6E6FD}{$\mathbf{0.008}^{\pm 0.003}$} \\
Run
& $113.727^{\pm 6.179}$
& \cellcolor[HTML]{E6E6FD}{$\mathbf{65.667}^{\pm 10.702}$}
& $74.405^{\pm 6.179}$
& \cellcolor[HTML]{E6E6FD}{$\mathbf{0.015}^{\pm 0.000}$}
& $0.0165^{\pm 0.0014}$
& $0.016^{\pm 0.001}$ \\
Jump
& $222.953^{\pm 10.702}$
& $188.001^{\pm 6.179}$
& \cellcolor[HTML]{E6E6FD}{$\mathbf{183.632}^{\pm 10.702}$}
& $0.022^{\pm 0.003}$
& $0.0228^{\pm 0.0021}$
& \cellcolor[HTML]{E6E6FD}{$\mathbf{0.020}^{\pm 0.003}$} \\
Spinkick
& $135.572^{\pm 6.179}$
& \cellcolor[HTML]{E6E6FD}{$\mathbf{87.512}^{\pm 6.179}$}
& $109.358^{\pm 6.179}$
& $0.031^{\pm 0.000}$
& $0.0326^{\pm 0.0013}$
& \cellcolor[HTML]{E6E6FD}{$\mathbf{0.030}^{\pm 0.001}$} \\
Cartwheel
& $183.632^{\pm 0.000}$
& $157.417^{\pm 10.702}$
& \cellcolor[HTML]{E6E6FD}{$\mathbf{144.310}^{\pm 18.536}$}
& $0.031^{\pm 0.000}$
& \cellcolor[HTML]{E6E6FD}{$\mathbf{0.0275}^{\pm 0.0026}$}
& $0.028^{\pm 0.003}$ \\
Backflip
& Failed
& $179.263^{\pm 6.179}$
& \cellcolor[HTML]{E6E6FD}{$\mathbf{157.417}^{\pm 10.702}$}
& Failed
& $0.0461^{\pm 0.0017}$
& \cellcolor[HTML]{E6E6FD}{$\mathbf{0.045}^{\pm 0.001}$} \\
Sideflip
& $336.549^{\pm 62.708}$
& $166.156^{\pm 12.358}$
& \cellcolor[HTML]{E6E6FD}{$\mathbf{161.787}^{\pm 6.179}$}
& $0.051^{\pm 0.002}$
& $0.0508^{\pm 0.0026}$
& \cellcolor[HTML]{E6E6FD}{$\mathbf{0.049}^{\pm 0.001}$} \\
Speed Vault
& $388.978^{\pm 49.430}$
& $122.465^{\pm 12.358}$
& \cellcolor[HTML]{E6E6FD}{$\mathbf{118.096}^{\pm 0.000}$}
& $0.024^{\pm 0.001}$
& $0.0218^{\pm 0.0012}$
& \cellcolor[HTML]{E6E6FD}{$\mathbf{0.021}^{\pm 0.003}$} \\
Double Kong
& Failed
& $284.120^{\pm 6.179}$
& \cellcolor[HTML]{E6E6FD}{$\mathbf{205.477}^{\pm 6.179}$}
& Failed
& $0.0291^{\pm 0.0016}$
& \cellcolor[HTML]{E6E6FD}{$\mathbf{0.028}^{\pm 0.001}$} \\

\midrule
\multicolumn{7}{c}{\textbf{General Motion Tracking Across Episode Lengths}} \\
\midrule
\textbf{Episode}
& \multicolumn{3}{c}{\textbf{Position Error [m] ($\downarrow$)}}
& \multicolumn{3}{c}{\textbf{Rotation Error [rad] ($\downarrow$)}} \\
\cmidrule(lr){2-4}
\cmidrule(lr){5-7}
\textbf{Length [s]}
& \textbf{ADD}
& \textbf{ADD w/ PDF-HR}
& \textbf{ADD w/ PFM-HR}
& \textbf{ADD}
& \textbf{ADD w/ PDF-HR}
& \textbf{ADD w/ PFM-HR} \\
\midrule
10
& $0.022^{\pm 0.001}$
& $0.023^{\pm 0.002}$
& \cellcolor[HTML]{E6E6FD}{$\mathbf{0.021}^{\pm 0.001}$}
& $0.121^{\pm 0.007}$
& $0.129^{\pm 0.011}$
& \cellcolor[HTML]{E6E6FD}{$\mathbf{0.119}^{\pm 0.009}$} \\
20
& $0.023^{\pm 0.002}$
& $0.023^{\pm 0.003}$
& \cellcolor[HTML]{E6E6FD}{$\mathbf{0.020}^{\pm 0.002}$}
& $0.123^{\pm 0.007}$
& $0.127^{\pm 0.008}$
& \cellcolor[HTML]{E6E6FD}{$\mathbf{0.117}^{\pm 0.008}$} \\
30
& $0.022^{\pm 0.001}$
& $0.022^{\pm 0.002}$
& \cellcolor[HTML]{E6E6FD}{$\mathbf{0.020}^{\pm 0.002}$}
& $0.122^{\pm 0.008}$
& $0.125^{\pm 0.007}$
& \cellcolor[HTML]{E6E6FD}{$\mathbf{0.117}^{\pm 0.007}$} \\
\bottomrule
\end{tabular}
\caption{Quantitative comparison of sample efficiency and tracking accuracy.
The upper block reports single-trajectory tracking results, where samples denote the number of samples required to first reach $SR \geq 80\%$ and position errors are reported in meters.
The lower block reports general motion tracking across different episode lengths using joint position and rotation errors.
We benchmark our method against ADD~\cite{zhang2025physics} and ADD w/ PDF-HR~\cite{gu2026hr}.
Results are reported as mean with standard deviation in superscript across three independent random seeds.
A motion is marked as ``Failed'' if all three trials fail to converge within $6000$ training iterations.
Highlighted entries indicate the best mean result for each skill and metric.}
\label{table:single_motion_tracking_exp}
\end{table*}

%% file: Table/ablation.tex
\begin{table*}[!t]
\centering
\small
\setlength{\tabcolsep}{5pt}
\renewcommand{\arraystretch}{1.05}
\begin{tabular}{lccc ccc}
\specialrule{.16em}{.08em}{.08em}
\textbf{Skill}
& \multicolumn{3}{c}{\textbf{Samples ($SR \geq 80\%$) [M] ($\downarrow$)}}
& \multicolumn{3}{c}{\textbf{Position Error [m] ($\downarrow$)}} \\
\cmidrule(lr){2-4}
\cmidrule(lr){5-7}
& \textbf{$\boldsymbol{v}$-Pred}
& \textbf{FM-Recon}
& \textbf{Ours}
& \textbf{$\boldsymbol{v}$-Pred}
& \textbf{FM-Recon}
& \textbf{Ours} \\
\midrule

Walk
& $70.036^{\pm 6.179}$
& \cellcolor[HTML]{E6E6FD}{$\mathbf{56.929}^{\pm 6.179}$}
& $65.667^{\pm 0.000}$
& \cellcolor[HTML]{E6E6FD}{$\mathbf{0.008}^{\pm 0.003}$}
& \cellcolor[HTML]{E6E6FD}{$\mathbf{0.008}^{\pm 0.001}$}
& \cellcolor[HTML]{E6E6FD}{$\mathbf{0.008}^{\pm 0.003}$} \\

Run
& $83.143^{\pm 6.179}$
& \cellcolor[HTML]{E6E6FD}{$\mathbf{74.405}^{\pm 16.348}$}
& \cellcolor[HTML]{E6E6FD}{$\mathbf{74.405}^{\pm 6.179}$}
& $0.017^{\pm 0.000}$
& $0.018^{\pm 0.001}$
& \cellcolor[HTML]{E6E6FD}{$\mathbf{0.016}^{\pm 0.001}$} \\

Backflip
& $170.525^{\pm 18.536}$
& $170.525^{\pm 18.536}$
& \cellcolor[HTML]{E6E6FD}{$\mathbf{157.417}^{\pm 10.702}$}
& $0.047^{\pm 0.001}$
& $0.046^{\pm 0.000}$
& \cellcolor[HTML]{E6E6FD}{$\mathbf{0.045}^{\pm 0.001}$} \\

Sideflip
& $166.156^{\pm 6.179}$
& \cellcolor[HTML]{E6E6FD}{$\mathbf{161.787}^{\pm 12.358}$}
& \cellcolor[HTML]{E6E6FD}{$\mathbf{161.787}^{\pm 6.179}$}
& $0.051^{\pm 0.001}$
& $0.051^{\pm 0.003}$
& \cellcolor[HTML]{E6E6FD}{$\mathbf{0.049}^{\pm 0.001}$} \\

\specialrule{.16em}{.08em}{.08em}
\end{tabular}
\caption{Ablation studies of the Flow Matching prediction
parameterization and reward paradigm.
We compare PFM-HR against two variants on
single-trajectory tracking: $\boldsymbol{v}$-Pred replaces the
$\boldsymbol{x}$-prediction parameterization with
$\boldsymbol{v}$-prediction, while FM-Recon uses the same pretrained Flow
Matching prior but replaces the PGS reward with an SDS-style reward. }
\label{table:combined_ablation}
\end{table*}

%% file: Supplementary_Material_arxiv.tex
\section{Denoiser-Induced Pose Geometry}
\label{sec:supp_derivation}

This section establishes the connection between the population-optimal
clean-pose predictor and the conditional covariance of clean poses, and then
derives the Pose Geometry Score (PGS). Fix $t\in(0,1-\delta]$ and let $\sigma_t=1-t$, where $\delta>0$ is a small constant (we use $\delta=10^{-3}$) that clamps the velocity denominator near $t=1$. We assume that the clean-pose distribution has a finite
second moment and satisfies the regularity conditions required to exchange
differentiation and integration below. Let $D$ denote the dimension of the
pose representation. In this work, we use $D = N_J = 29 $ for the Unitree G1 humanoid robot, matching the normalized pose representation of the main paper.

\subsection{Population-Optimal Clean-Pose Predictor}

Let $\boldsymbol{x}\sim p_{\mathrm{data}}$ and
$\boldsymbol{\epsilon}\sim\mathcal{N}(\mathbf{0},\mathbf{I})$ be independent,
where $p_{\mathrm{data}}$ denotes the clean-pose distribution (a probability
measure on $\mathbb{R}^D$).
Under the linear Flow Matching path, with $\boldsymbol{x}$ and
$\boldsymbol{\epsilon}$ held fixed along the path,
\begin{equation}
    \boldsymbol{z}_t
    =t\boldsymbol{x}+\sigma_t\boldsymbol{\epsilon},
    \boldsymbol{v}
    =\frac{d\boldsymbol{z}_t}{dt}
    =\boldsymbol{x}-\boldsymbol{\epsilon}
    =\frac{\boldsymbol{x}-\boldsymbol{z}_t}{\sigma_t}.
    \label{eq:supp_fm_path}
\end{equation}
A clean-pose predictor $\hat{\boldsymbol{x}}_\phi$ induces the velocity field
\begin{equation}
    \boldsymbol{v}_\phi(\boldsymbol{z}_t,t)
    =
    \frac{
        \hat{\boldsymbol{x}}_\phi(\boldsymbol{z}_t,t)-\boldsymbol{z}_t
    }{\sigma_t}.
    \label{eq:supp_pred_velocity}
\end{equation}
Therefore, at a fixed timestep $t$, the Flow Matching objective can be written
as
\begin{equation}
    \mathcal{L}_{\mathrm{FM},t}(\phi)
    =
    \frac{1}{\sigma_t^2}
    \mathbb{E}\!\left[
        \left\|
        \boldsymbol{x}-\hat{\boldsymbol{x}}_\phi(\boldsymbol{z}_t,t)
        \right\|_2^2
    \right].
    \label{eq:supp_fm_regression}
\end{equation}

The derivation below considers the regime $t\leq 1-\delta$, which is
also the regime used for PGS evaluation. In this regime, the denominator
clamp in the implemented velocity parameterization is inactive and
the velocity residual is exactly proportional to the clean-pose
regression residual.
Because $\sigma_t^{-2}$ depends only on $t$, it does not alter the pointwise
population minimizer at that timestep. Thus, over the unrestricted class of
square-integrable predictors, a population-optimal clean-pose predictor is
\begin{equation}
    \mathbf{F}_t^\star(\boldsymbol{z})
    =
    \mathbb{E}\!\left[
        \boldsymbol{x}\mid\boldsymbol{z}_t=\boldsymbol{z}
    \right],
    \label{eq:supp_conditional_mean}
\end{equation}
defined $p_t(\boldsymbol{z})$-almost everywhere, where
$p_t(\boldsymbol{z})=\int p_t(\boldsymbol{z}\mid\boldsymbol{x})
\,\mu_{\mathrm{data}}(d\boldsymbol{x})$ is the marginal density of
$\boldsymbol{z}_t$.

\subsection{Jacobian--Conditional Covariance Identity}

The Gaussian likelihood induced by the path in
Eq.~\eqref{eq:supp_fm_path}, together with its score, is
\begin{align}
    p_t(\boldsymbol{z}\mid\boldsymbol{x})
    &=
    \frac{1}{(2\pi\sigma_t^2)^{D/2}}
    \exp\!\left(
        -\frac{\|\boldsymbol{z}-t\boldsymbol{x}\|_2^2}
              {2\sigma_t^2}
    \right),
    \label{eq:supp_likelihood}\\
    \nabla_{\boldsymbol{z}}\log p_t(\boldsymbol{z}\mid\boldsymbol{x})
    &=
    -\frac{\boldsymbol{z}-t\boldsymbol{x}}{\sigma_t^2}.
    \label{eq:supp_likelihood_score}
\end{align}
Let $\mu_{\mathrm{data}}$ denote the clean-pose probability measure, i.e.,
the measure underlying $p_{\mathrm{data}}$ introduced above. The
conditional mean can be expressed as
\begin{equation}
    \mathbf{F}_t^\star(\boldsymbol{z})
    =
    \frac{
        \int \boldsymbol{x}\,p_t(\boldsymbol{z}\mid\boldsymbol{x})
        \,\mu_{\mathrm{data}}(d\boldsymbol{x})
    }{
        \int p_t(\boldsymbol{z}\mid\boldsymbol{x})
        \,\mu_{\mathrm{data}}(d\boldsymbol{x})
    }.
\end{equation}
Differentiating this ratio gives, for coordinates $i,j$,
\begin{align}
    \frac{\partial F_{t,i}^\star(\boldsymbol{z})}{\partial z_j}
    &=
    \operatorname{Cov}\!\left[
        x_i,
        \frac{\partial}{\partial z_j}
        \log p_t(\boldsymbol{z}\mid\boldsymbol{x})
        \,\middle|\,
        \boldsymbol{z}_t=\boldsymbol{z}
    \right]
    \nonumber\\
    &=
    \frac{t}{\sigma_t^2}
    \operatorname{Cov}\!\left[
        x_i,x_j
        \,\middle|\,
        \boldsymbol{z}_t=\boldsymbol{z}
    \right].
    \label{eq:supp_component_jacobian}
\end{align}
Consequently,
\begin{equation}
    \mathbf{J}_t^\star(\boldsymbol{z})
    :=
    \nabla_{\boldsymbol{z}}\mathbf{F}_t^\star(\boldsymbol{z})
    =
    \frac{t}{(1-t)^2}\mathbf{C}_t(\boldsymbol{z}),
    \label{eq:supp_jacobian_covariance}
\end{equation}
where  $ \mathbf{C}_t(\boldsymbol{z})
    :=
    \operatorname{Cov}\!\left[
        \boldsymbol{x}\mid\boldsymbol{z}_t=\boldsymbol{z}
    \right]$. This is the Jacobian--conditional covariance identity used in Eq.~(11) of the
main paper. In particular, the off-diagonal entries of
$\mathbf{C}_t(\boldsymbol{z})$ describe posterior co-variation between pose
coordinates at the queried noisy pose.

\subsection{Induced Positive-Semidefinite Geometry}

For a differentiable learned predictor $\mathbf{F}_\phi$, define

\begin{align}
    \mathbf{J}_\phi(\boldsymbol{z},t)
   & =
    \nabla_{\boldsymbol{z}}\mathbf{F}_\phi(\boldsymbol{z},t), \\
    \mathbf{G}_\phi(\boldsymbol{z},t)
    &=
    \mathbf{J}_\phi(\boldsymbol{z},t)^\top
    \mathbf{J}_\phi(\boldsymbol{z},t).
    \label{eq:supp_geometry}
\end{align}

For any $\mathbf{a}\in\mathbb{R}^{D}$,
\begin{equation}
    \mathbf{a}^\top
    \mathbf{G}_\phi(\boldsymbol{z},t)
    \mathbf{a}
    =
    \|\mathbf{J}_\phi(\boldsymbol{z},t)\mathbf{a}\|_2^2
    \geq 0.
    \label{eq:supp_psd_proof}
\end{equation}
Thus, $\mathbf{G}_\phi$ defines a positive-semidefinite pullback quadratic
form, possibly degenerate, which we refer to as the denoiser-induced local
pose geometry. This establishes Eq.~(12) of the main paper. At the population
optimum, Eq.~\eqref{eq:supp_jacobian_covariance} and the symmetry of the
conditional covariance give
\begin{equation}
    \mathbf{G}_t^\star(\boldsymbol{z})
    =
    \left(
        \frac{t}{(1-t)^2}
    \right)^2
    \mathbf{C}_t(\boldsymbol{z})^2.
    \label{eq:supp_population_geometry}
\end{equation}
where $\mathbf{C}_t(\boldsymbol{z})^2$ denotes the matrix square
$\mathbf{C}_t(\boldsymbol{z})\mathbf{C}_t(\boldsymbol{z})$.

\subsection{Pose Geometry Score}

At rollout step $k$, the finite-difference joint-coordinate velocity and its
stabilized direction are

\begin{align}
        \boldsymbol{\xi}_{\mathrm{env},k}
    &=
    \frac{\mathbf{q}_k-\mathbf{q}_{k-1}}{\Delta t}, \\ 
    \mathbf{d}_k
    &=
    \frac{\boldsymbol{\xi}_{\mathrm{env},k}}
    {\sqrt{
        \|\boldsymbol{\xi}_{\mathrm{env},k}\|_2^2
        +\eta_{\mathrm{num}}
    }}.
    \label{eq:supp_direction}
\end{align}

Here, $\eta_{\mathrm{num}}>0$ denotes a small numerical stabilization
constant (we use $\eta_{\mathrm{num}}=10^{-6}$; this is the
$\varepsilon_{\mathrm{num}}$ of the main paper) that prevents division by
near-zero values, and $\mathbf{q}_k$ denotes the normalized joint-coordinate
pose representation used in the main paper. For a corruption draw
$\boldsymbol{\epsilon}_k\sim\mathcal{N}(\mathbf{0},\mathbf{I})$, let
\begin{align}
        \widetilde{\mathbf{q}}_{k-1}
    &=
    t_{\mathrm{eval}}\mathbf{q}_{k-1}
    +(1-t_{\mathrm{eval}})\boldsymbol{\epsilon}_k, \\
    \mathbf{J}_{\phi,k}
    &=
    \mathbf{J}_\phi\!\left(
        \widetilde{\mathbf{q}}_{k-1},
        t_{\mathrm{eval}}
    \right),
\end{align}
where $t_{\mathrm{eval}}\leq 1-\delta$. The PGS is then
\begin{align}
    s_{\mathrm{PGS},k}
    &:=
    \|\mathbf{J}_{\phi,k}\mathbf{d}_k\|_2^2
    \nonumber\\
    &=
    \frac{
        \|\mathbf{J}_{\phi,k}
        \boldsymbol{\xi}_{\mathrm{env},k}\|_2^2
    }{
        \|\boldsymbol{\xi}_{\mathrm{env},k}\|_2^2
        +\eta_{\mathrm{num}}
    }.
    \label{eq:supp_pgs}
\end{align}
This recovers Eq.~(15) of the main paper. The product
$\mathbf{J}_{\phi,k}\mathbf{d}_k$ can be evaluated directly with one
Jacobian--vector product, without explicitly constructing either
$\mathbf{J}_{\phi,k}$ or $\mathbf{G}_{\phi,k}$.

To obtain the population interpretation, suppose that the learned Jacobian
agrees with the population-optimal Jacobian at the queried point, and define
\begin{equation}
    \mathbf{J}_{\phi,k}
    =
    \mathbf{J}_{t_{\mathrm{eval}}}^\star
    (\widetilde{\mathbf{q}}_{k-1}),
    \qquad
    \mathbf{C}_k
    =
    \mathbf{C}_{t_{\mathrm{eval}}}
    (\widetilde{\mathbf{q}}_{k-1}).
\end{equation}
Equation~\eqref{eq:supp_jacobian_covariance} then gives
\begin{equation}
    s_{\mathrm{PGS},k}^\star
    =
    \left(
        \frac{t_{\mathrm{eval}}}
             {(1-t_{\mathrm{eval}})^2}
    \right)^2
    \mathbf{d}_k^\top
    \mathbf{C}_k^2
    \mathbf{d}_k,
    \label{eq:supp_population_pgs}
\end{equation}
which is Eq.~(16) of the main paper.

Because $\mathbf{C}_k$ is symmetric and positive semidefinite, it admits the
eigendecomposition
\begin{equation}
    \mathbf{C}_k
    =
    \mathbf{U}_k
    \operatorname{diag}
    (\lambda_{k,1},\ldots,\lambda_{k,D})
    \mathbf{U}_k^\top,
    \qquad
    \lambda_{k,i}\geq 0.
\end{equation}
Writing the stabilized rollout direction in this eigenbasis as
\begin{equation}
    \mathbf{d}_k
    =
    \sum_{i=1}^{D}a_{k,i}\mathbf{u}_{k,i},
    \qquad
    \sum_{i=1}^{D}a_{k,i}^2
    =
    \|\mathbf{d}_k\|_2^2
    \leq 1,
\end{equation}
we obtain
\begin{equation}
    s_{\mathrm{PGS},k}^\star
    =
    \left(
        \frac{t_{\mathrm{eval}}}
             {(1-t_{\mathrm{eval}})^2}
    \right)^2
    \sum_{i=1}^{D}
    \lambda_{k,i}^2a_{k,i}^2.
    \label{eq:supp_eigen_pgs}
\end{equation}
Thus, at a fixed corruption timestep, the population-optimal PGS is large
when the stabilized rollout direction has substantial components along
posterior pose-variation modes with large conditional covariance eigenvalues.
This interpretation concerns the local co-variation structure of the marginal
pose distribution and does not require temporally ordered prior-training data.
For a learned predictor, the interpretation is approximate to the extent that
its input Jacobian agrees with the population-optimal Jacobian at the queried
point.

\section{Robustness of Online PGS Evaluation}                              
\label{sec:pgs_robustness}                                   
PGS is evaluated online using a stochastically corrupted pose query.          
Given a transition $(\mathbf q_i,\mathbf d_i)$, we corrupt the pose        
query by interpolating toward Gaussian noise at the evaluation
timestep $t_{\mathrm{eval}}$:
\begin{equation}
  \widetilde{\mathbf q}_i
  =
  t_{\mathrm{eval}}\mathbf q_i
  +(1-t_{\mathrm{eval}})\boldsymbol\epsilon_i, 
  \boldsymbol\epsilon_i
  \sim\mathcal N(\mathbf 0,\mathbf I),
  \label{eq:corrupted_query}
\end{equation}
and obtain the score
\begin{equation}
  s_{\mathrm{PGS}}(\mathbf q_i,\mathbf d_i;t_{\mathrm{eval}})
  =
  \left\|
  J_\phi\!\left(
  \widetilde{\mathbf q}_i,t_{\mathrm{eval}}
  \right)\mathbf d_i
  \right\|_2^2.
  \label{eq:pgs_score}
\end{equation}
We use the same frozen pose prior as in the main experiments and
evaluate on transitions sampled from 100 motions randomly selected
from the BONES-SEED dataset; all reported metrics are averaged over
these transitions. We note that these transitions are drawn from the pose
prior's training corpus; the study measures the internal consistency of the
learned PGS under pose-change perturbations rather than its generalization
to held-out poses. The remainder of this section is organized as
follows. We first select the evaluation timestep by measuring how well
PGS discriminates coordinated joint changes from structurally
perturbed ones across timesteps. We then verify that, at the selected
timestep, the single-sample estimate remains consistent under
stochastic corruption. Finally, we show that PGS is insensitive to the
overall magnitude of a pose change while remaining sensitive to the
inter-joint coordination pattern.

\subsection{Effect of the Evaluation Timestep}
\label{sec:timestep_structural_discriminability}

The evaluation timestep determines the corruption scale at which the
frozen pose prior is queried, and it may consequently affect the
ability of PGS to distinguish coordinated joint changes from
structurally perturbed ones. To study this, we measure the structural
discriminability of PGS across
\begin{equation}
t_{\mathrm{eval}}
\in
\{0.4,0.5,0.6,0.7,0.75,0.8,0.9\}.
\end{equation}

For every transition, we retain the original pose $\mathbf q_i$ and
the joint-change direction $\mathbf d_i$, and construct
perturbed change directions that preserve the pose query. To measure
how the structural discrimination varies with the amount of
disruption, we sweep the severity of each perturbation:

\begin{table*}[!ht]
  \centering
  \small
  \caption{Robustness and computational cost of PGS under
  stochastic corruption samples. The 128-sample estimate is used
  as the high-accuracy offline reference. }
  \label{tab:pgs_noise_robustness}
  \setlength{\tabcolsep}{6pt}
  \begin{tabular}{cccccc}
      \toprule
      $K$
      & Rank corr. $\uparrow$
      & Percentile MAE $\downarrow$
      & Region agree. $\uparrow$
      & Reward MAE $\downarrow$
      & Latency (ms) $\downarrow$ \\
      \midrule
      1
      & 0.976 & 0.013 & 97.9\% & 0.010 & 0.750 \\
      2
      & 0.981 & 0.011 & 98.4\% & 0.007 & 1.308 \\
      4
      & 0.990 & 0.009 & 98.9\% & 0.004 & 2.847 \\
      8
      & 0.996 & 0.005 & 99.3\% & 0.002 & 5.991 \\
      Ref. ($K=128$)
      & 1.000 & 0.000 & 100.0\% & 0.000 & 97.112 \\
      \bottomrule
  \end{tabular}
\end{table*}

\noindent
\textbf{Angular perturbation.} The change direction is rotated by
$\theta\in\{15^\circ,30^\circ,45^\circ,60^\circ,75^\circ,90^\circ\}$
within the plane spanned by $\mathbf d_i$ and a unit vector
$\mathbf n_{\perp,i}$ sampled uniformly from the orthogonal complement of
$\mathbf d_i$:
\begin{equation}
\widetilde{\mathbf d}_i(\theta)
=
\cos(\theta)\mathbf d_i
+
\sin(\theta)\mathbf n_{\perp,i}.
\label{eq:angular_perturbation}
\end{equation}
Which rotates the change direction while approximately preserving its norm
(the stabilized direction $\mathbf d_i$ deviates from unit norm only by the
small stabilization term of Eq.~\eqref{eq:supp_direction}).

\noindent
\textbf{Joint permutation.} A random subset comprising a fraction
$\rho\in\{25\%,50\%,75\%,100\%\}$ of the joint-coordinate blocks is
selected and permuted, disrupting the relative change pattern while
preserving the set of local changes.

\noindent
\textbf{Independent sign flip.} A random subset comprising a fraction
$\rho\in\{25\%,50\%,75\%,100\%\}$ of the joint-coordinate blocks is
selected, and the change directions of the selected blocks are
reversed. For $\rho<100\%$, this disrupts the relative change pattern; at
$\rho=100\%$ it is a global sign flip, which leaves the quadratic PGS
invariant at a fixed query pose.

For each transition, the original and perturbed directions are
evaluated using the corrupted pose query of
Eq.~\eqref{eq:corrupted_query} and the Gaussian noise sample. The corrupted
query and the Gaussian noise sample are held fixed across the original and
perturbed evaluations, so their score difference is caused only by the
change in the inter-joint structure.

\noindent
\textbf{Timestep-specific calibration.}
Raw PGS values cannot be compared directly across evaluation
timesteps because the scale of the denoiser Jacobian depends on
$t_{\mathrm{eval}}$. We therefore independently recompute the
reference PGS distribution and empirical CDF
$\widehat{F}_{\tau,t}$ for every timestep, evaluating each reference PGS
with a single corruption sample from Eq.~\eqref{eq:corrupted_query},
consistent with the $K=1$ online estimator.

For the original transition, we compute
\begin{equation}
  u_i(t)
  =
  \widehat{F}_{\tau,t}
  \left(
      s_{\mathrm{PGS}}
      (\mathbf q_i,\mathbf d_i;t)
  \right),
\end{equation}
and for a perturbed direction $\widetilde{\mathbf d}_i$,
\begin{equation}
  \widetilde{u}_i(t)
  =
  \widehat{F}_{\tau,t}
  \left(
      s_{\mathrm{PGS}}
      (\mathbf q_i,\widetilde{\mathbf d}_i;t)
  \right).
\end{equation}

We define the structural discrimination score for perturbation
$\mathcal P$ at severity level $s$ as the average decrease in the
calibrated percentile:
\begin{equation}
  D_{\mathcal P,s}(t)
  =
  \frac{1}{N}
  \sum_{i=1}^{N}
  \left[
      u_i(t)-\widetilde{u}_i^{\mathcal P,s}(t)
  \right].
\end{equation}
Because each perturbation is swept over a set of severity levels
$\mathcal S_{\mathcal P}$ (the rotation angle $\theta$ for the angular
perturbation and the joint fraction $\rho$ for joint permutation and
sign flip), we report the severity-averaged score
\begin{equation}
  D_{\mathcal P}(t)
  =
  \frac{1}{|\mathcal S_{\mathcal P}|}
  \sum_{s\in\mathcal S_{\mathcal P}}
  D_{\mathcal P,s}(t).
\end{equation}
A larger positive value indicates that PGS assigns substantially lower
reference-relative scores after the joint-change structure is
disrupted and therefore provides stronger structural discrimination.
Values close to zero indicate that PGS responds similarly to the
original and perturbed change patterns.

\begin{table}[t]
    \centering
    \small
    \caption{Structural discrimination of PGS at different evaluation
    timesteps. Each entry reports the decrease in reference-relative
    percentile after perturbation, averaged over the severity levels
    of each perturbation. Higher values indicate stronger
    discrimination. The reference PGS distribution is recalibrated
    independently for every timestep.}
    \label{tab:timestep_structural_discriminability}
    \setlength{\tabcolsep}{4.5pt}
    \begin{tabular}{ccccc}
        \toprule
        $t_{\mathrm{eval}}$
        & Angular
        & Permutation
        & Sign flip
        & Average \\
        \midrule
        0.40 & 0.061 & 0.050 & 0.019 & 0.043 \\
        0.50 & 0.089 & 0.076 & 0.028 & 0.064 \\
        0.60 & 0.121 & 0.105 & 0.040 & 0.089 \\
        0.70 & 0.159 & 0.135 & 0.053 & 0.116 \\
        \cellcolor[HTML]{E6E6FD}{\textbf{0.75}}
        & \cellcolor[HTML]{E6E6FD}{\textbf{0.169}}
        & \cellcolor[HTML]{E6E6FD}{\textbf{0.146}}
        & \cellcolor[HTML]{E6E6FD}{\textbf{0.057}}
        & \cellcolor[HTML]{E6E6FD}{\textbf{0.124}} \\
        0.80 & 0.164 & 0.138 & 0.054 & 0.119 \\
        0.90 & 0.106 & 0.086 & 0.033 & 0.075 \\
        \bottomrule
    \end{tabular}
\end{table}

\begin{table*}[t]
  \centering
  \small
  \caption{Robustness of PGS to magnitude and joint-change-pattern
  perturbations. $\Delta$Percentile and $\Delta r^P$ are measured
  relative to the unperturbed transition.}
  \label{tab:pgs_perturbation_robustness}
  \setlength{\tabcolsep}{7pt}
  \begin{tabular}{clcccc}
      \toprule
      \textbf{Perturbation}
      & \textbf{Severity}
      & \textbf{Normalized PGS}
      & $\boldsymbol{\Delta}$\textbf{Percentile}
      & $\boldsymbol{\Delta r^P}$
      & \textbf{Region agreement} \\
      \midrule
      None
      & --
      & 1.000 & 0.000 & 0.000 & 100.0\% \\
      \midrule

      \multirow{4}{*}{\shortstack[c]{Magnitude\\scaling}}
      & $c=0.25$
      & 0.984 & $-0.004$ & $-0.001$ & 99.1\% \\
      & $c=0.5$
      & 0.995 & $-0.001$ & $-0.000$ & 99.7\% \\
      & $c=2$
      & 1.001 & $+0.000$ & $+0.000$ & 99.8\% \\
      & $c=4$
      & 1.002 & $+0.001$ & $+0.000$ & 99.6\% \\
      \midrule

      \multirow{6}{*}{\shortstack[c]{Angular\\perturbation}}
      & $\theta=15^\circ$
      & 0.928 & $-0.018$ & $-0.006$ & 96.1\% \\
      & $\theta=30^\circ$
      & 0.794 & $-0.057$ & $-0.021$ & 89.4\% \\
      & $\theta=45^\circ$
      & 0.623 & $-0.126$ & $-0.058$ & 78.6\% \\
      & $\theta=60^\circ$
      & 0.441 & $-0.214$ & $-0.118$ & 65.2\% \\
      & $\theta=75^\circ$  
      & 0.350 & $-0.280$ & $-0.160$ & 57.5\% \\
      & $\theta=90^\circ$
      & 0.268 & $-0.318$ & $-0.205$ & 50.7\% \\
      \midrule

      \multirow{4}{*}{\shortstack[c]{Joint\\permutation}}
      & 25\% joints
      & 0.842 & $-0.043$ & $-0.015$ & 91.8\% \\
      & 50\% joints
      & 0.654 & $-0.112$ & $-0.047$ & 81.0\% \\
      & 75\% joints  
      
      & 0.510 & $-0.180$ & $-0.100$ & 70.3\% \\
      & 100\% joints
      & 0.372 & $-0.251$ & $-0.151$ & 59.5\% \\
      \midrule

      \multirow{4}{*}{\shortstack[c]{Independent\\sign flip}}
      & 25\% joints
      & 0.878 & $-0.032$ & $-0.010$ & 93.4\% \\
      & 50\% joints
      & 0.712 & $-0.091$ & $-0.035$ & 84.8\% \\
      & 75\% joints 
      & 0.660 & $-0.105$ & $-0.040$ & 83.0\% \\
      & 100\% joints
      & 1.000 & $0.000$ & $0.000$ & 100.0\% \\
      \bottomrule
      
  \end{tabular}
\end{table*}

\begin{table}[h]
    \centering
    \small
    \setlength{\tabcolsep}{5pt}
    \renewcommand{\arraystretch}{1.10}
    \caption{Architecture and training configuration of
    the PFM-HR pose prior.}
    \label{tab:supp_network_architecture}
    \begin{tabular}{lc}
        \toprule
        \textbf{Component} & \textbf{Configuration} \\
        \midrule
        Input/output dimension & $N_J$=29 \\
        Backbone & Residual MLP \\
        Number of residual blocks & 10 \\
        Hidden dimension & 1024 \\
        Timestep embedding & Sinusoidal embedding + MLP \\
        Timestep conditioning & adaLN-Zero \\
        Normalization & RMSNorm \\
        Activation & SiLU \\
        Prediction target & Clean pose  \\
        Training loss & Flow Matching velocity loss \\
        \bottomrule
    \end{tabular}
\end{table}

\begin{table*}[h]
    \centering
    \small
    \caption{PPO implementation details. The hyperparameters follow
    the official MimicKit ADD configuration.}
    \label{tab:ppo_hyperparameters}
    \setlength{\tabcolsep}{5pt}
    \begin{tabular}{lclc}
        \toprule
        \textbf{Hyperparameter} & \textbf{Value}
        & \textbf{Hyperparameter} & \textbf{Value} \\
        \midrule
        Parallel environments       & 4,096
        & Rollout length             & 32 \\
        Samples per iteration        & 131,072
        & Discount factor $\gamma$   & 0.99 \\
        TD($\lambda$) / GAE parameter & 0.95
        & PPO clipping ratio         & 0.2 \\
        Normalized advantage clipping & 4.0
        & Actor optimization epochs  & 5 \\
        Critic optimization epochs   & 2
        & Actor minibatch size       & 16,384 \\
        Critic minibatch size        & 8,192
        & Actor learning rate        & $1\times10^{-4}$ \\
        Critic learning rate         & $1\times10^{-4}$
        & Optimizer                  & SGD \\
        Actor network                & $[1024,1024]$
        & Critic network             & $[1024,1024]$ \\
        Action distribution          & Fixed Gaussian
        & Action standard deviation  & 0.05 \\
        Actor output init.\ scale    & 0.01
        & Action-bound coefficient   & 10.0 \\
        Entropy coefficient          & 0.0
        & Action regularization      & 0.0 \\
        Mixed-precision training     & Disabled
        & Normalizer samples         & $10^8$ \\
        \bottomrule
    \end{tabular}
\end{table*}

Table~\ref{tab:timestep_structural_discriminability} shows an
inverted-U-shaped relationship between the evaluation timestep and
the structural discriminability of PGS. The average discrimination
score increases from $0.043$ at $t_{\mathrm{eval}}=0.4$ to $0.124$
at $t_{\mathrm{eval}}=0.75$, and subsequently decreases to $0.075$
at $t_{\mathrm{eval}}=0.9$. A similar trend is observed consistently
across angular perturbation, joint permutation, and independent sign
flip.

The neighboring timesteps $0.7$ and $0.8$ also obtain relatively
strong average scores of $0.116$ and $0.119$, respectively, indicating
that PGS remains effective within an intermediate timestep range
rather than depending on a narrowly tuned value. Among the evaluated
settings, $t_{\mathrm{eval}}=0.75$ provides the strongest average
response to disrupted joint-change structures, which is consistent with the
timestep selected by the validation sweep reported in the main paper.

\subsection{Stability under Stochastic Corruption}
\label{sec:pgs_noise_stability}

At the selected evaluation timestep $t_{\mathrm{eval}}=0.75$, we first
verify that the online estimate remains consistent under stochastic
corruption. For a fixed transition $(\mathbf q_i,\mathbf d_i)$, we
draw $K$ independent corrupted queries from Eq.~\eqref{eq:corrupted_query}
and obtain $K$ score samples
$s_i^{(j)}=s_{\mathrm{PGS}}(\mathbf q_i,\mathbf d_i;t_{\mathrm{eval}})$,
$j=1,\ldots,K$, one per corrupted query. The estimate obtained using
$K$ independent corruption samples is
\begin{equation}
  \widehat{s}_i^{(K)}
  =
  \frac{1}{K}\sum_{j=1}^{K}s_i^{(j)}.
\end{equation}
We use an independently sampled $K_{\mathrm{ref}}=128$ estimate as the
high-accuracy reference and compare it with
$K\in\{1,2,4,8\}$. Importantly, the proposed method uses $K=1$ during
policy training, drawing a fresh corruption sample at every rollout step, and
therefore requires only one JVP per rollout step. We report four metrics. The Spearman rank correlation measures how
faithfully the ordering of the $K$-sample estimates matches the
reference ordering across transitions. The percentile MAE is the mean
absolute error between the reference-relative percentiles obtained
from the $K$-sample and reference estimates, and the reward MAE is the
corresponding mean absolute error of the resulting reward multiplier.
Finally, we measure the calibration-region agreement: each
reference-relative percentile $u$ is mapped to one of three reward
regions by
\begin{equation}
  \mathcal B(u)
  =
  \begin{cases}
      0, & u < p_{\mathrm{bad}},\\
      1, & p_{\mathrm{bad}}\leq u < p_{\mathrm{good}},\\
      2, & u \geq p_{\mathrm{good}},
  \end{cases}
  \label{eq:region_binning}
\end{equation}
and the agreement is the fraction of transitions for which the
$K$-sample and reference estimates fall into the same region. The same
binning is reused for the perturbation study below. Latency is measured for
  one online PGS evaluation step using a batch
size of 4096 on an NVIDIA GeForce RTX 4090 GPU.

\noindent
\textbf{Results.}
As shown in Table~\ref{tab:pgs_noise_robustness}, the single-sample
estimator achieves a Spearman rank correlation of $0.976$ and a
calibration-region agreement of $97.9\%$ with the 128-sample
reference, while adding only $0.75$ ms per evaluation step, over two
orders of magnitude cheaper than the $97.1$ ms reference estimate.
Increasing the number of corruption samples provides only marginal
improvements, supporting the use of a single JVP as a stable and efficient online
PGS estimator during policy training.

\subsection{Robustness to Pose-Change Perturbations}
\label{sec:pgs_perturbation}

At the selected evaluation timestep $t_{\mathrm{eval}}=0.75$, we
examine whether PGS distinguishes the overall magnitude of a
joint-coordinate change from its inter-joint coordination pattern.

For magnitude scaling, denoting by
$\boldsymbol{\xi}_i=(\mathbf{q}_i-\mathbf{q}_{i-1})/\Delta t$ the
finite-difference velocity of transition $i$, we construct
\begin{equation}
  \boldsymbol\xi_i^{(c)}
  =
  c\,\boldsymbol\xi_i,
  \qquad
  c\in\{0.25,0.5,1,2,4\},
\end{equation}
and obtain the corresponding normalized direction as
\begin{equation}
  \mathbf d_i^{(c)}
  =
  \frac{\boldsymbol\xi_i^{(c)}}
  {\sqrt{\|\boldsymbol\xi_i^{(c)}\|_2^2
  +\eta_{\mathrm{num}}}}.
\end{equation}
Since PGS depends primarily on the normalized joint-change direction,
its response should remain approximately invariant to $c$, except
when the change magnitude becomes comparable to the numerical
stabilization term.

To assess the sensitivity to the inter-joint coordination pattern, we
evaluate the perturbation family of
Section~\ref{sec:timestep_structural_discriminability} across
severities: the angular rotation of Eq.~\eqref{eq:angular_perturbation}
at
$\theta\in\{15^\circ,30^\circ,45^\circ,60^\circ,75^\circ,90^\circ\}$, and joint
permutation and independent sign flips applied to
$\rho\in\{25\%,50\%,75\%,100\%\}$ of the joint-coordinate blocks.
Table~\ref{tab:pgs_perturbation_robustness} reports all of these
severities.
The corrupted pose query (Eq.~\eqref{eq:corrupted_query}) and sampled
noise are held fixed before and after each perturbation, ensuring that
the observed difference is caused only by the joint-change direction.

Let $s_i$, $u_i$, and $r_i^P$ denote the PGS, reference-relative
percentile, and reward multiplier of the original transition,
respectively. Their perturbed counterparts are denoted by
$\widetilde{s}_i$, $\widetilde{u}_i$, and
$\widetilde{r}_i^P$. We evaluate each perturbation using four metrics,
averaged over $N$ transitions.

First, the normalized PGS measures the fraction of the original score
retained after perturbation (with the stabilization constant
$\eta_{\mathrm{num}}$ added to both scores to keep the ratio stable near
zero):
\begin{equation}
  R_{\mathrm{PGS}}
  =
  \frac{1}{N}
  \sum_{i=1}^{N}
  \frac{
      \widetilde{s}_i+\eta_{\mathrm{num}}
  }{
      s_i+\eta_{\mathrm{num}}
  }.
\end{equation}
A value close to one indicates that the PGS response is preserved,
whereas a value below one indicates a reduced alignment with the
geometry encoded by the pose prior.

Second, the percentile change measures the signed displacement in the
reference-calibrated PGS distribution:
\begin{equation}
  \Delta\mathrm{Percentile}
  =
  \frac{1}{N}
  \sum_{i=1}^{N}
  \left(\widetilde{u}_i-u_i\right).
\end{equation}
Negative values indicate that the perturbed joint-change patterns are
assigned lower reference-relative percentiles.

Third, the reward change measures the corresponding effect on the PGS
reward multiplier:
\begin{equation}
  \Delta r^P
  =
  \frac{1}{N}
  \sum_{i=1}^{N}
  \left(\widetilde{r}_i^P-r_i^P\right).
\end{equation}
A negative value indicates that the perturbation leads to stronger
downweighting of the tracking reward.

Fourth, the region agreement measures whether a transition remains in
the same reward-calibration region after perturbation, using the
binning $\mathcal B(\cdot)$ of Eq.~\eqref{eq:region_binning}:
\begin{equation}
  A_{\mathrm{region}}
  =
  \frac{1}{N}
  \sum_{i=1}^{N}
  \mathbf{1}
  \left[
      \mathcal B(\widetilde{u}_i)
      =
      \mathcal B(u_i)
  \right].
\end{equation}
Magnitude scaling is expected to produce
$R_{\mathrm{PGS}}\approx1$, small percentile and reward changes, and
high region agreement. In contrast, perturbations that progressively
disrupt the inter-joint coordination pattern are expected to reduce
$R_{\mathrm{PGS}}$, decrease the calibrated percentile and reward
multiplier, and lower the region agreement.

\begin{table*}[h]
    \centering
    \small
    \setlength{\tabcolsep}{4.5pt}
    \renewcommand{\arraystretch}{1.10}
    \caption{
       Rotational tracking error on the single-trajectory
        tracking tasks.
        We compare vanilla ADD, ADD with PDF-HR, and ADD with
        PFM-HR. Results are reported in radians as mean
        $\pm$ one standard deviation across three independent
        random seeds. A method is marked as ``Failed'' when all
        three trials fail to converge within 6000 training
        iterations.
    }
    \label{tab:supp_rotation_error}
    \begin{tabular}{lccc}
        \toprule
        \textbf{Skill}
        & \textbf{ADD}
        & \textbf{ADD w/ PDF-HR}
        & \textbf{ADD w/ PFM-HR} \\
        \midrule
        Walk
        & $\cellcolor[HTML]{E6E6FD}{\mathbf{0.038 \pm 0.001}}$
        & $0.042 \pm 0.002$
        & $0.039 \pm 0.000$ \\

        Run
        & $\cellcolor[HTML]{E6E6FD}{\mathbf{0.063 \pm 0.001}}$
        & $0.070 \pm 0.001$
        & $0.069 \pm 0.003$ \\

        Jump
        & $0.108 \pm 0.004$
        & $0.103 \pm 0.003$
        & $\cellcolor[HTML]{E6E6FD}{\mathbf{0.101 \pm 0.005}}$ \\

        Spinkick
        & $0.150 \pm 0.002$
        & $0.161 \pm 0.004$
        & $\cellcolor[HTML]{E6E6FD}{\mathbf{0.147 \pm 0.002}}$ \\

        Cartwheel
        & $0.129 \pm 0.007$
        & $0.126 \pm 0.010$
        & $\cellcolor[HTML]{E6E6FD}{\mathbf{0.125 \pm 0.005}}$ \\

        Backflip
        & Failed
        & $0.205 \pm 0.004$
        & $\cellcolor[HTML]{E6E6FD}{\mathbf{0.203 \pm 0.003}}$ \\

        Sideflip
        & $0.241 \pm 0.003$
        & $0.242 \pm 0.005$
        & $\cellcolor[HTML]{E6E6FD}{\mathbf{0.238 \pm 0.001}}$ \\

        SpeedVault
        & $0.118 \pm 0.003$
        & $\cellcolor[HTML]{E6E6FD}{\mathbf{0.116 \pm 0.001}}$
        & $\cellcolor[HTML]{E6E6FD}{\mathbf{0.116 \pm 0.000}}$ \\

        DoubleKong
        & Failed
        & $0.157 \pm 0.003$
        & $\cellcolor[HTML]{E6E6FD}{\mathbf{0.154 \pm 0.002}}$ \\
        \bottomrule
    \end{tabular}
\end{table*}

\begin{figure*}[h]
\centering
\includegraphics[width=0.99\linewidth]{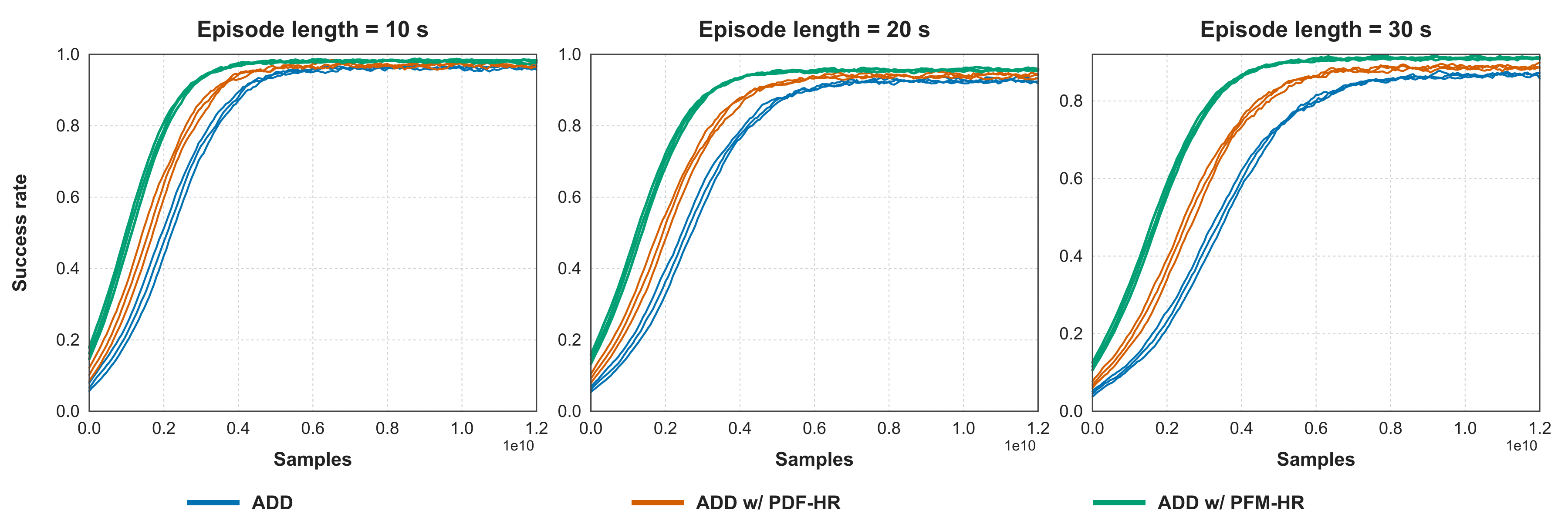}

\caption{
Training curves for general-motion tracking on the 34-sequence LaFAN1
subset. The three panels report success rates for episode horizons of
10, 20, and 30 seconds, respectively. We compare ADD (blue), ADD with
PDF-HR (orange), and ADD with PFM-HR (green). The horizontal axis
denotes the number of training samples, and each curve corresponds to
one of three independent random seeds. Higher success rates indicate
that a larger fraction of the reference motion is tracked
continuously.
}
\label{fig:general_motion_training_curves}

\end{figure*}

\begin{figure*}[h]
\centering
\includegraphics[width=0.99\linewidth]{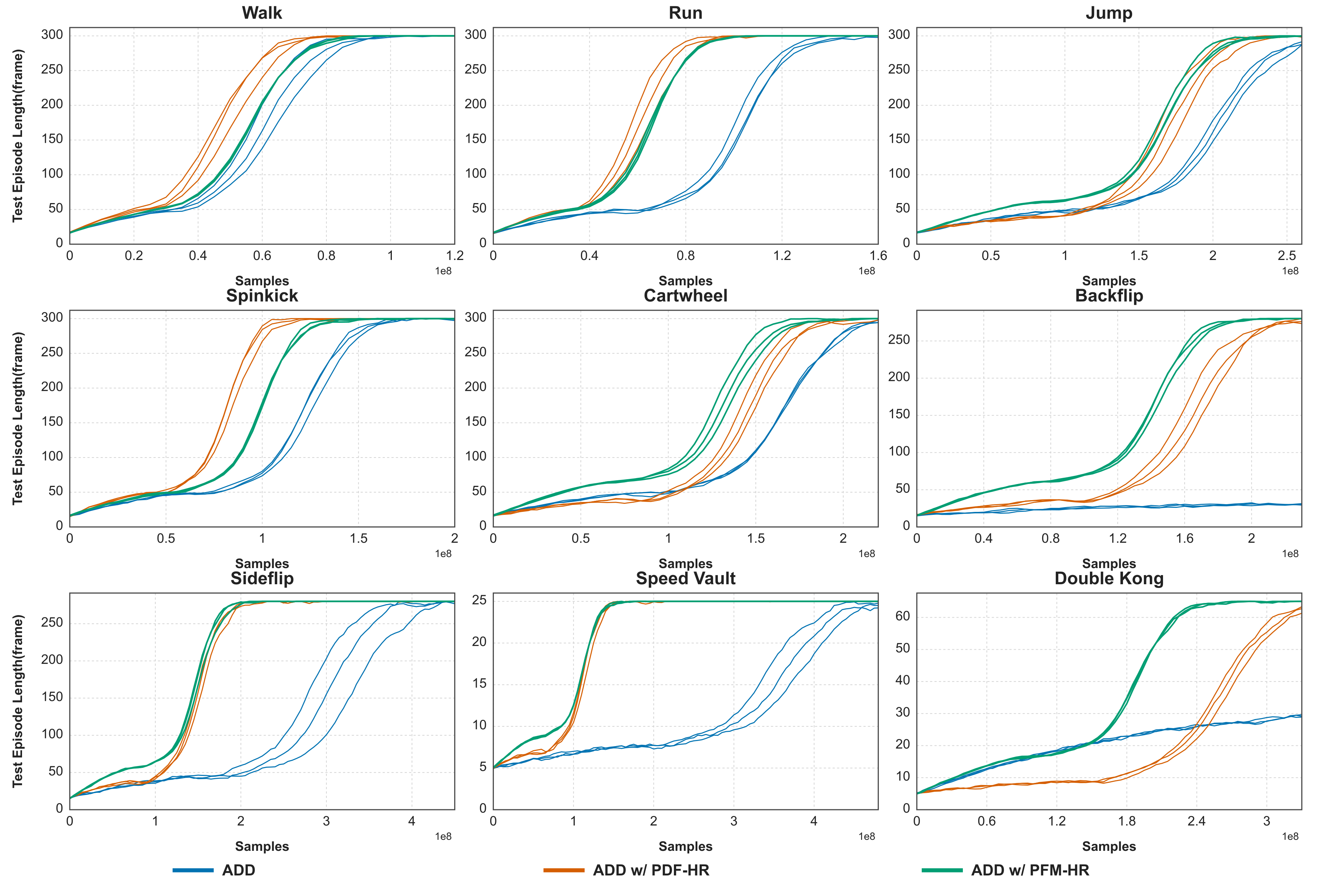}
\caption{
Training curves on nine single-motion tracking tasks.
We compare ADD (blue), ADD with PDF-HR (orange), and ADD with
PFM-HR (green). The horizontal axis denotes the number of environment
samples, and the vertical axis reports the test episode length in
frames, where a larger value indicates that the policy tracks the
reference motion for a longer duration before termination. Each curve
corresponds to one of three independent random seeds. The maximum
episode length differs across tasks according to the duration of the
corresponding reference motion.
}
\label{fig:single_motion_training_curves}

\end{figure*}

\noindent
\textbf{Results.}
Table~\ref{tab:pgs_perturbation_robustness} shows that PGS is nearly
invariant to the overall magnitude of a pose change. Across scaling
factors from $0.25$ to $4$, the normalized score remains within
$1.6\%$ of the unperturbed value, and the calibration-region
agreement remains above $99\%$. In contrast, PGS decreases
consistently as the joint-change pattern is perturbed. At an angular
deviation of $90^\circ$, only $26.8\%$ of the original score is
retained, and fully permuting the joint changes reduces the
normalized score to $0.372$. Independently flipping the directions of
a random subset of joints also lowers the score, to $0.878$ and
$0.712$ at $25\%$ and $50\%$ of the joints, respectively, whereas a
global sign flip of all joint directions leaves the score unchanged
($1.000$), consistent with the sign-invariance of the quadratic PGS
at a fixed query pose. These results suggest that PGS is robust to
nuisance variations while remaining sensitive to the inter-joint
coordination patterns.

\section{Implementation Details}

\subsection{Pose Flow Matching Network Architecture}
\label{sec:supp_network_architecture}

The pose prior is pretrained on the 60M-pose BONES-SEED corpus using
the same normalized $N_J=29$ dimensional representation as in the main
paper. We use the linear Flow Matching path
$\boldsymbol{z}_t=t\boldsymbol{x}+(1-t)\boldsymbol\epsilon$ and sample $t$ from
the logit-normal distribution used in our implementation. The network
is trained with the clean-pose parameterization and the induced
velocity-space loss. The prior is pretrained independently of all
downstream policies and is frozen before reinforcement learning.

PFM-HR follows the clean-data prediction formulation of
JiT, adapted from image inputs to vector-valued
humanoid poses. Since each pose is represented by an
$N_J$-dimensional joint-coordinate vector, we use a residual MLP
instead of a patch-based Transformer. The network consists of an
input projection, 10 residual MLP blocks with a hidden dimension
of 1024, and a linear output projection. The Flow Matching
timestep is encoded using sinusoidal features followed by a
two-layer MLP and is injected into every residual block through
adaLN-Zero conditioning. The output dimension is $N_J$, allowing
the network to directly predict the clean normalized pose. Following JiT, we use
$\boldsymbol{x}$-prediction with a velocity-space loss. Given a clean pose
$\boldsymbol{x}$, Gaussian noise
$\boldsymbol{\epsilon}\sim\mathcal{N}(\mathbf{0},\mathbf{I})$,
and timestep $t$, the corrupted pose is
\begin{equation}
    \boldsymbol{z}_t
    =
    t\boldsymbol{x}+(1-t)\boldsymbol{\epsilon}.
\end{equation}
The network directly predicts the clean pose,
\begin{equation}
    \hat{\boldsymbol{x}}_\phi
    =
    F_\phi(\boldsymbol{z}_t,t),
\end{equation}
and the corresponding velocity prediction is
\begin{equation}
    \boldsymbol{v}_\phi(\boldsymbol{z}_t,t)
    =
    \frac{
        \hat{\boldsymbol{x}}_\phi-\boldsymbol{z}_t
    }{
        \max(1-t,\delta)
    }.
\end{equation}
Using the target velocity
$\boldsymbol{}{v}=\boldsymbol{x}-\boldsymbol{\epsilon}$, we optimize
\begin{equation}
    \mathcal{L}_{\mathrm{FM}}(\phi)
    =
    \mathbb{E}_{t,\boldsymbol{x},\boldsymbol{\epsilon}}
    \left[
        \left\|
            \boldsymbol{v}
            -
            \boldsymbol{v}_\phi(\boldsymbol{z}_t,t)
        \right\|_2^2
    \right].
\end{equation}
Thus, the model output remains a clean-pose prediction, while the
training error is evaluated in the induced Flow Matching velocity
space. The resulting clean-pose map is subsequently frozen and
used to compute the Jacobian--vector products required by PGS.

\subsection{RL Training Details}

We implement all policies based on the ADD framework in MimicKit
 and use the Isaac Gym backend with 4,096
parallel environments. Unless otherwise specified, the PPO
hyperparameters and actor--critic architecture follow the official
ADD configuration as shown in Table~\ref{tab:ppo_hyperparameters}. The pretrained prior in PFM-HR is kept frozen
throughout policy learning. All discriminator-related hyperparameters, including the discriminator
architecture, learning rate, replay-buffer size, replay-sampling count,
gradient penalty, and discriminator reward scale, follow the official
MimicKit ADD configuration without modification.

The actor and critic are two-layer fully connected networks with
1,024 hidden units per layer. MimicKit specifies the actor and critic
minibatch sizes as multiples of the number of parallel environments.
Therefore, the corresponding minibatch sizes are
$4\times4096=16{,}384$ and $2\times4096=8{,}192$, respectively.
All compared methods use the same training budget and environment
configuration. We evaluate each policy every 100 training iterations
using 4,096 episodes and report the mean and standard deviation over
three seeds.

\section{Additional Experimental Results}
\label{sec:supp_additional_results}

\subsection{Rotation Error on Single-Trajectory Tasks}
\label{sec:supp_rotation_results}

Table~\ref{tab:supp_rotation_error} reports the rotational
tracking errors for all nine single-trajectory tracking tasks.
These results complement the positional tracking errors reported
in Table~2 of the main paper. We use the same evaluation protocol
and report the mean and one standard deviation across three
independent random seeds. PFM-HR obtains the lowest mean rotational error on six of the
nine tasks and ties for the lowest error on SpeedVault. The
vanilla ADD baseline performs best on the two basic locomotion
tasks, Walk and Run. In contrast, PFM-HR performs best on the
more dynamic tasks, including Jump, Spinkick, Cartwheel,
Backflip, Sideflip, and Double Kong. In particular, PFM-HR
maintains low rotational errors on Backflip and Double Kong,
for which vanilla ADD fails.

\subsection{Training Curves on Single-motion Tracking}                        

Figure~\ref{fig:single_motion_training_curves} presents the complete          
training curves for the nine single-motion tracking tasks. Both pose-prior methods generally improve
sample efficiency over vanilla ADD, yet the magnitude of the benefit
depends strongly on the complexity of the reference motion. On dynamic
and acrobatic skills such as Jump, Cartwheel, Backflip, Sideflip,
SpeedVault, and Double Kong, PFM-HR converges particularly rapidly,
reaching the full reference length well before vanilla ADD. Its
advantage is most pronounced on Backflip and Double Kong, where
vanilla ADD fails to reach the full reference length within the
training budget, whereas PFM-HR consistently solves the motions across
all three seeds. On the comparatively simpler Walk, Run, and Spinkick
tasks, where joint trajectories are more predictable and rely less on
tightly coordinated multi-joint timing, PDF-HR converges slightly
faster than or comparably to PFM-HR, although both methods remain
substantially more sample-efficient than ADD. 

\subsection{Training Curves on General Motion Tracking}

Figure~\ref{fig:general_motion_training_curves} shows the success-rate
curves for general-motion tracking under three episode horizons (10,
20, and 30 seconds). Longer horizons require the policy to track the
reference over many more timesteps and to recover from error
accumulation, making them a substantially harder test of reward
quality. PFM-HR consistently shifts the learning curves to the left,
reaching high success rates with fewer training samples than both
vanilla ADD and ADD with PDF-HR. This advantage is observed across all
three horizons and becomes more evident as the episode horizon
increases, suggesting that the benefit of geometry-aware reward
modulation grows with the amount of accumulated drift the policy must
correct. PDF-HR also improves sample efficiency over vanilla ADD, but
its gains are smaller than those of PFM-HR, indicating that the richer
inter-joint geometry information used by PFM-HR translates into a
clearer learning signal. In addition to faster convergence, PFM-HR
achieves the highest final success rate across all three horizons,
showing that PGS-based reward modulation improves both learning speed
and long-horizon tracking reliability rather than trading one off
against the other.